%% file: main.tex
\documentclass[10pt,twocolumn,letterpaper]{article}

\usepackage[pagenumbers]{cvpr} % To force page numbers, e.g. for an arXiv version

\input{preamble}

\makeatletter
\def\@fnsymbol#1{\ifcase#1\or $\dagger$\or $*$\or $\ddagger$\or $\mathsection$\or $\mathparagraph$\or $\|$\fi}
\makeatother
 
\def\paperID{11502} % *** Enter the Paper ID here
\def\confName{CVPR}
\def\confYear{2025}

\title{\name: Memory-Guided Diffusion for Expressive Talking Video Generation}

\author{Longtao Zheng$^{2*}$\thanks{Work done during the internship at Skywork AI.}~, Yifan Zhang$^{1}$\thanks{Authors contributed equally.}~\thanks{Project Lead.}~, Hanzhong Guo, Jiachun Pan$^{1}$, Zhenxiong Tan$^{3}$, \\Jiahao Lu$^{3\dagger}$, Chuanxin Tang$^{1}$, Bo An$^{1,2}$, Shuicheng Yan$^{1}$\\  
$^{1}$Skywork AI,
$^{2}$Nanyang Technological University,
$^{3}$National University of Singapore\\ 
{\tt\small longtao001@e.ntu.edu.sg, \{yifan.zhang7,shuicheng.yan\}@kunlun-inc.com}\\
Project Page: \url{https://memoavatar.github.io}
}

\begin{document}
\maketitle
\begin{abstract}

Recent advances in video diffusion models have unlocked new potential for realistic audio-driven talking video generation. However, achieving seamless audio-lip synchronization, maintaining long-term identity consistency, and producing natural, audio-aligned expressions in generated talking videos remain significant challenges. To address these challenges, we propose \textbf{M}emory-guided \textbf{EMO}tion-aware diffusion (MEMO), an end-to-end audio-driven portrait animation approach to generate identity-consistent and expressive talking videos. Our approach is built around two key modules: (1) a memory-guided temporal module, which enhances long-term identity consistency and motion smoothness by developing memory states to store information from a longer past context to guide temporal modeling via linear attention; and (2) an emotion-aware audio module, which replaces traditional cross attention with multi-modal attention to enhance audio-video interaction, while detecting emotions from audio to refine facial expressions via emotion adaptive layer norm. Extensive quantitative and qualitative results demonstrate that MEMO generates more realistic talking videos across diverse image and audio types, outperforming state-of-the-art methods in overall quality, audio-lip synchronization, identity consistency, and expression-emotion alignment.

\end{abstract}

\section{Introduction}

Audio-driven talking video generation~\cite{prajwal2020lip,tian2024emo,xu2024vasa} has gained significant attention due to its broad impact on areas like virtual avatars, digital content creation, and real-time communication, offering transformative possibilities in entertainment, education, and e-commerce. However, compared to text-to-video generation~\cite{guo2023animatediff,rombach2022high,ramesh2022hierarchical} or image-to-video generation~\cite{blattmann2023stable}, audio-driven talking video generation presents unique challenges. It requires not only generating synchronized lip movements and realistic head motions from audio, but also preserving long-term identity consistency of the reference image and producing natural expressions that align with the emotional tone of audio. Balancing these demands while ensuring generalization across diverse driving audio and reference images makes this task especially challenging.

\begin{figure}[t]
%\vspace{-0.1in}
\centering
\includegraphics[width=0.45\textwidth]{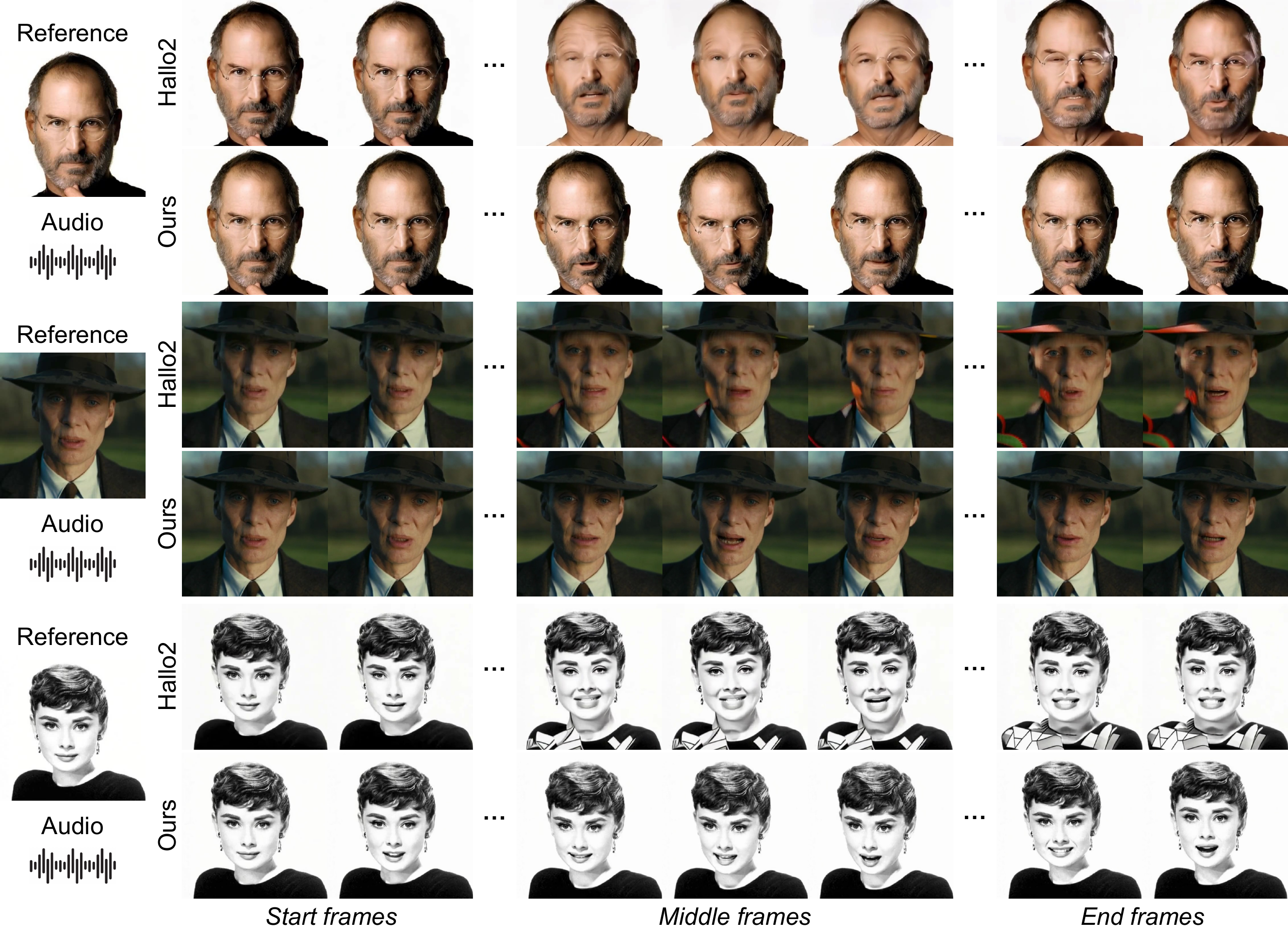}
\vspace{-0.1in}
\caption{Our \name generates talking videos with improved identity consistency, audio-lip alignment, and motion smoothness. In contrast, existing diffusion methods (\emph{e.g.}, Hallo2~\cite{cui2024hallo2}) are prone to temporal error accumulation during autoregressive generation, especially when the last 2-4 generated frames used as temporal conditions contain artifacts, leading to inconsistent identity. Please refer to the supplementary material for video demos.} 
\label{fig:demo}
\vspace{-0.1in}
\end{figure}

Recent advances in video diffusion models~\cite{tian2024emo,xu2024hallo,chen2024echomimic,cui2024hallo2} have enabled more realistic audio-driven talking video generation. Most of these methods use cross attention to incorporate audio to guide video generation, and typically condition on past 2-4 generated frames for autoregressive generation to improve motion smoothness~\cite{tian2024emo,xu2024hallo}. Additionally, some incorporate a single, human-defined emotion label for the whole video to specify the emotion of the generated video~\cite{xu2024vasa,tan2024flowvqtalker}. However, these approaches face challenges with audio-lip synchronization, maintaining long-term identity consistency, and achieving natural expressions aligned with the audio. Specifically, cross attention relies on fixed audio features and limits audio-video interaction, while conditioning on a limited number of past frames can lead to temporal error accumulation, especially when those frames contain artifacts (cf.~Figure~\ref{fig:demo}). Moreover, using a fixed emotion label for the whole video can result in facial expressions that fail to capture the dynamic emotional shifts inherent in audio. As a result, these methods struggle with audio-lip synchronization, expression-audio alignment, and long-term identity preservation. 

\begin{figure*}[t]
\vspace{-0.2in}
\centering
\includegraphics[width=0.77\textwidth]{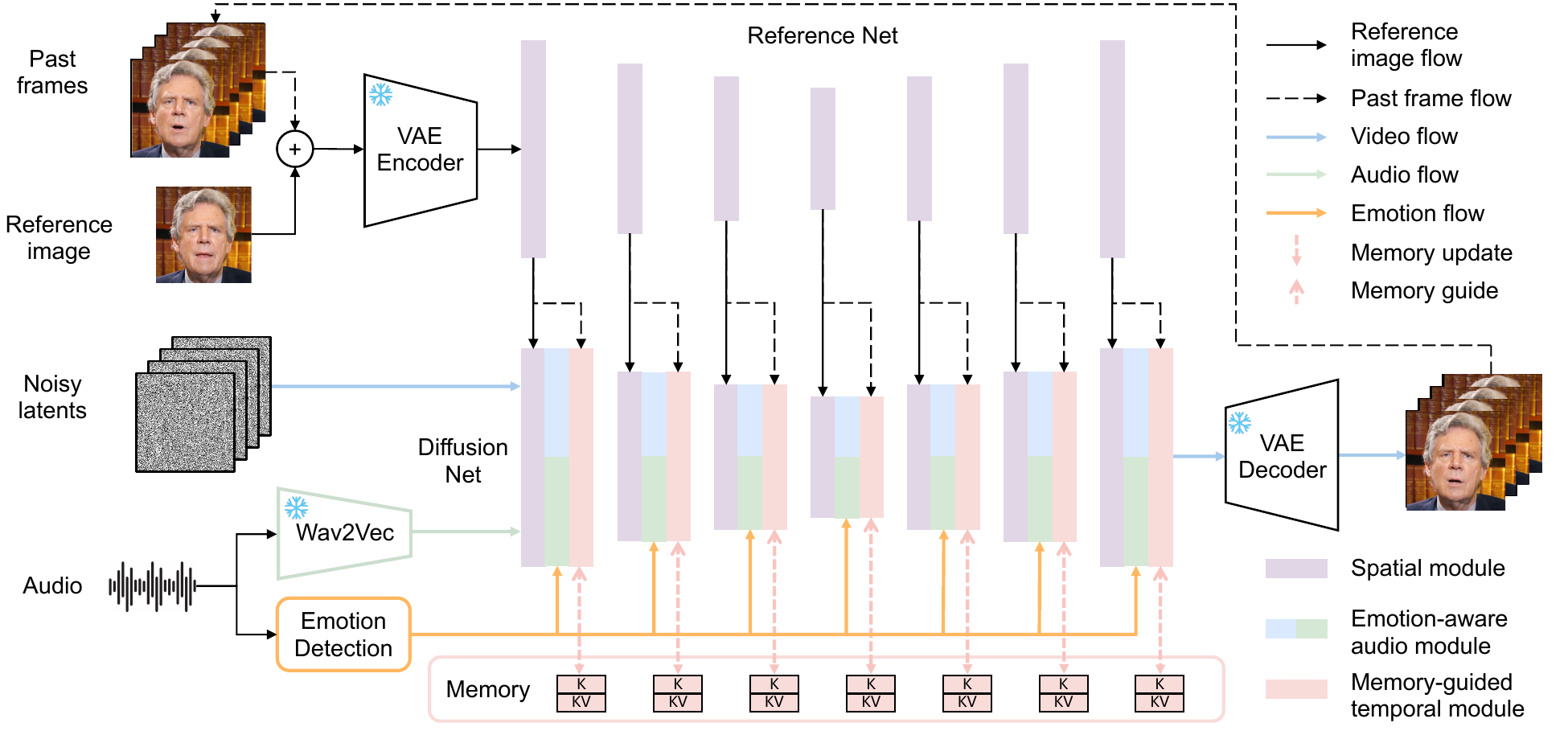}
\vspace{-0.1in}
\caption{Overview of \name, which is structured with a Reference Net and a Diffusion Net. The core innovations of \name reside in two key modules within the Diffusion Net: the \textbf{memory-guided temporal module}  and the \textbf{emotion-aware audio module}. These modules work in tandem to deliver enhanced audio-video synchronization, sustained identity consistency, and more natural expression generation.} 
\label{fig:method}
\vspace{-0.15in}
\end{figure*}

In this paper, we propose \textbf{M}emory-guided \textbf{EMO}tion-aware diffusion (\name), an end-to-end audio-driven portrait animation approach. As shown in Figure~\ref{fig:method}, \name is built around two key modules: (1) a memory-guided temporal module and (2) an emotion-aware audio module. To ensure consistent facial identity and smooth transitions across long-duration videos, \name develops a memory-guided temporal module  (cf. Section~\ref{sec:temporalmodule}) that maintains memory states across longer previously generated frames. This allows the model to use long-term motion information to guide temporal modeling through linear attention, resulting in more coherent facial movements and mitigating the error accumulation issue that may occur in existing diffusion methods (cf. Figure~\ref{fig:demo}). Moreover, to improve audio-lip synchronization and align facial expressions with the audio emotion, \name introduces an emotion-aware audio module (cf. Section~\ref{sec:audiomodule}). This module replaces the traditional cross-attention audio module in previous diffusion methods with a more dynamic multi-modal attention mechanism, enabling better interaction between audio and video during the diffusion process. Meanwhile, by dynamically detecting emotion cues from the audio at the video subsegment level, this module helps to subtly refine facial expressions via the emotion adaptive layer norm, enabling the generation of expressive talking videos.

Extensive quantitative results and human evaluations demonstrate that our approach consistently outperforms state-of-the-art methods in overall quality, audio-lip synchronization, expression-audio alignment, identity consistency, and motion smoothness  (cf. Table~\ref{tab:combined_distribution} and Figure~\ref{fig:human_study}). Additionally, diverse qualitative results highlight \name's strong generalization across various types of audio, images, languages, and head poses (cf. Figures~\ref{fig:special_audio_demo}-\ref{fig:sideface_demo}), further showcasing the effectiveness of our method. Lastly, ablation studies further validate the distinct contributions of the memory-guided temporal module (cf. Figure~\ref{fig:ablate_f_human}), which enhances long-term identity consistency and motion smoothness, and the emotion-aware audio module (cf. Figure~\ref{fig:mm_dif_human}-\ref{fig:emotion_demo}), which significantly improves audio-lip alignment and expression naturalness. 

\section{Related Work}

Audio-driven talking video generation aims to synthesize realistic and synchronized talking videos given driving audio and a reference face image. Early approaches only focused on learning audio-lip mapping while keeping other facial attributes static~\cite{suwajanakorn2017synthesizing,chen2018lip,prajwal2020lip,cheng2022videoretalking,yin2022styleheat}. These methods, however, cannot capture comprehensive facial expressions and natural head movements. To resolve this, later research used intermediate motion representations (\emph{e.g.}, landmark coordinates, 3D facial mesh, and 3D morphable models) and decomposed the generation process into two stages, \emph{i.e.}, audio to motion and motion to video~\cite{zhou2020makelttalk,sun2023vividtalk,zhang2023sadtalker,wang2024v,chen2024echomimic,wei2024aniportrait}. However, they often generate inaccurate intermediate representations from audio, which restricts the expressiveness and realism of the resulting videos.

Recent end-to-end methods, like EMO~\cite{tian2024emo} and Hallo~\cite{xu2024hallo}, can generate vivid portrait videos by fine-tuning pre-trained diffusion models~\cite{rombach2022high}. However, they need specific modules (\emph{e.g.}, face locator) to constrain head stability, which makes it impossible for the model to achieve naturally large head motions. Similar issues exist in the methods that learned a specific face latent space~\cite{he2023gaia,ma2023dreamtalk,zhang2023dream,xu2024vasa,cui2024hallo2}. In contrast, our work does not depend on any facial inductive biases, which unlocks the possibilities for generating more expressive head motions in talking videos. Moreover, most of these methods use 2-4 past frames~\cite{tian2024emo,xu2024hallo,cui2024hallo2}  as temporal conditions for autoregressive generation of long videos. However, such limited frame history can result in error accumulation over time when artifacts appear in the past  2-4 frames. Very recently,  Loopy~\cite{jiang2024loopy} increased the number of past frames to reduce this dependence, and used a temporal segment module to model cross-clip relationships. However, it still uses limited past frames, whereas our proposed memory-guided linear attention module allows utilizing possibly all past frames to provide more comprehensive temporal guidance, thus mitigating error accumulation and enhancing long-term identity consistency. 
Besides, unlike previous diffusion-based methods~\cite{tian2024emo,xu2024hallo,cui2024hallo2} that used a cross-attention mechanism to integrate audio features, our method enhances the audio-lip synchronization and expression-audio alignment based on a newly developed emotion-aware multi-modal diffusion.
More related studies of diffusion models are provided in Appendix~\ref{app:related}.

\section{Problem and Preliminaries}

\paragraph{Problem statement.} Given a reference image and audio as inputs, audio-driven talking video generation~\cite{prajwal2020lip,tian2024emo} aims to output a vivid video that closely aligns with the input audio and authentically replicates real human speech and facial movements. This task is challenging because it requires seamless audio-lip synchronization, realistic head movements, long-term identity consistency, and natural expressions that align with audio. Most existing diffusion-based approaches~\cite{tian2024emo,xu2024hallo,chen2024echomimic} struggle with issues such as error accumulation, inconsistent identity preservation over time, limited audio-lip synchronization, unnatural expressions, and poor generalization.

\noindent \textbf{Latent diffusion models and rectified flow loss.} 
Our method is built upon the Latent Diffusion Model (LDM)~\cite{rombach2022high}, a framework designed to efficiently learn generative processes in a lower-dimensional latent space rather than directly operating on pixel space. During training, LDM first employs a pre-trained encoder $ \mathcal{E}(\cdot) $ to map high-dimensional images into a compressed latent space, producing latent features $ z_0 = \mathcal{E}(I) $. Then, following the principles of Denoising Diffusion Probabilistic Models (DDPM)~\cite{ho2020denoising}, Gaussian noise $ \epsilon $ is progressively added to the latent features over $ t $ discrete timesteps, resulting in noisy latent features $ z_t = \sqrt{\alpha_t} z_0 + \sqrt{1 - \alpha_t} \epsilon $, where $ \alpha_t $ is a variance schedule controlling how much noise is added. The diffusion model is then trained to reverse this noise-adding process by taking the noisy latent representation $ z_t $ as input and predicting the added noise $ \epsilon $. The objective function for training can be expressed as: $\mathcal{L} = \mathbb{E}_{z_t, c, \epsilon \sim \mathcal{N}(0,1), t} [ \| \epsilon - \epsilon_\theta (z_t, t, c) \|_2^2 ]$, where  $ \epsilon_\theta $ represents the noise prediction made by the U-Net network,  and  $ c $ represents conditioning information such as audio, or motion frames in the context of talking video generation. 

Recently, Stable Diffusion 3 (SD3)~\cite{esser2024scaling} refines this process by incorporating rectified flow loss~\cite{liu2023flow}, which modifies the traditional DDPM objective to: 
\begin{align}
\label{eqn:flow}
    \mathcal{L} = \mathbb{E}_{z_t, c, \epsilon \sim \mathcal{N}(0,1), t}[\lambda(t)\|\epsilon - \epsilon_\theta (z_t, t, c)\|_2^2],
\end{align}
where $\lambda(t) = 1/(1-t)^2$ and $z_t $ is reparameterized using linear combination as $ z_t = (1-t)z_0 + t\epsilon $. This formulation leads to both better training stability and more efficient inference. In light of these advantages, we adopt the rectified flow loss from SD3 in our training.

\section{Method}

As illustrated in Figure~\ref{fig:method}, \name is an end-to-end audio-driven diffusion model for generating identity-consistent and expressive talking videos.  Similar to previous diffusion-based approaches~\cite{tian2024emo,xu2024hallo}, \name is built around two main components: a Reference Net and a Diffusion Net. The main contributions of \name lie in two key modules within the Diffusion Net: the \textbf{memory-guided temporal module} (cf. Section~\ref{sec:temporalmodule}), and the \textbf{emotion-aware audio module} (cf. Section~\ref{sec:audiomodule}), which work together to achieve superior audio-video synchronization, long-term identity consistency, and natural expression generation. In addition, \name introduces a new data processing pipeline  (cf. Section~\ref{sec:datapipeline}) for acquiring high-quality talking head videos, along with a decomposed training strategy (cf. Section~\ref{sec:trainingstrategy}) to optimize diffusion model training.

\subsection{Memory-Guided Temporal Module}\label{sec:temporalmodule}

Most existing diffusion methods~\cite{tian2024emo,xu2024hallo,chen2024echomimic} generate talking videos in an autoregressive manner by segmenting the audio into clips of 12-16 frames and using the past 2-4 generated frames to condition the generation of the next video clip. They concatenate the past frame features with the current noisy latent features along the temporal dimension and apply temporal self-attention to model sequential information. While this approach can model short-term dependencies, it often struggles with maintaining consistency over longer sequences, even with the use of Reference Net. If artifacts are introduced in the past 2-4 conditioned frames, these errors tend to accumulate as generation progresses, leading to visual distortions that degrade both video quality and identity consistency (\eg, Hallo2~\cite{cui2024hallo2} in Figure~\ref{fig:demo}).

Motivated by the idea that leveraging a more complete memory of motion information, rather than relying solely on the most recent 2-4 frames, can provide richer guidance for enhancing identity consistency and motion smoothness, we propose a memory-guided temporal module. The key of this module is memory-guided linear attention, which is designed to improve temporal coherence and maintain consistent facial identity. 

\paragraph{Linear attention for temporal modeling.} Previous approaches use self-attention~\cite{tian2024emo,jiang2024loopy} to capture temporal information across frames. However, self-attention requires storing all key-value pairs, leading to increasing GPU memory overhead as the number of past frames grows, making it impractical to use longer motion information. To address this limitation, we replace self-attention with linear attention~\cite{katharopoulos2020transformers} and include a memory update mechanism into linear attention to model long-term temporal information efficiently. Denoting query as $Q$, key as $K$, and value as $V$, the output of linear attention for $i$-th frame is computed as:
\begin{align}
\text{out}_i = \frac{\phi(Q_i)^\top \big(\sum_{j=1}^f \phi(K_j) V_j^\top\big)}{\phi(Q_i)^\top \sum_{j=1}^f \phi(K_j)},
\end{align}
where $f$ is the frame number and $\phi$ is an activation function (we use softmax in this work).

\begin{figure}[t] 
\centering
\includegraphics[width=0.35\textwidth]{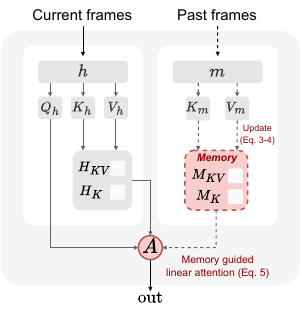}
\vspace{-0.1in}
\caption{Memory-guided temporal module.}
\label{fig:temporal_module}
\vspace{-0.2in}
\end{figure}

\paragraph{Memory update mechanism with history decay.} To incorporate motion information from a longer past context to guide video generation, we develop a memory update mechanism. Specifically, let the latent features of past frames as \( m \in \mathbb{R}^{f \times d} \) and the latent features of current frames as \( h \in \mathbb{R}^{f \times d} \), where \( d \) is the dimension of latent features. As shown in Figure~\ref{fig:temporal_module}, linear attention processes these latent features via learnable matrices, which transform them into queries (\( Q_h \)), keys (\( K_h, K_m \)), and values (\( V_h, V_m \)). 

To memorize motion information, we define the memory states for the past $f$ frames as two matrices: \( M^{f}_{KV} = \sum_{i=1}^{f} \gamma^i \phi( K_{m,i}) V_{m,i}^\top \) and \( M^{f}_K = \sum_{i=1}^{f} \gamma^i \phi(K_{m,i}) \), which occupy constant GPU memory irrespective of $f$. Here, \( \gamma \) is a decay factor \( (0 < \gamma < 1) \) that modulates the influence of past frames, with more recent frames exerting greater impact, reflected through the exponentiation by \( i \). After each generation of $f$ frames, we update the memory \( M^f \) by incorporating information from these newly generated frames. 
In formal, the memory update when adding the latest $a$ frames to the memory with $b$ past frames is:
\begin{align}
M^{a+b}_{KV} &\leftarrow \gamma^a M^{b}_{KV} + \textstyle \sum_{j=1}^{a} \gamma^{j} \phi(K_{h,j}) V_{h,j}^\top, \\
M^{a+b}_K &\leftarrow \gamma^a M^{b}_K + \textstyle \sum_{j=1}^{a} \gamma^{j} \phi(K_{h,j}).
\end{align}
Here, the decay scheme is critical, as a unified positional encoding across different video subsegments is infeasible. Instead, we use causal memory decay to provide implicit positional encoding, which enables more effective memory updates to capture long-term dependencies. During training, we set the temporal context to 16 past frames; at inference, this memory decay scheme allows the model to naturally extend memory updates to longer temporal contexts.

\paragraph{Memory-guided linear attention.} When generating the current video clip, we use the memory to guide temporal modeling. Let $H_{KV}=\phi(K_h)V_h^\top$ and $H_K = \phi(K_h)$. 
The output of the memory-guided temporal module is:
\begin{equation}
\text{out} = \frac{
\phi(Q_h)^\top \left(
H_{KV} + 
M_{KV}
\right) 
}{
\phi(Q_h)^\top \left(
H_K  + 
M_K
\right) 
}.
\end{equation}
This enables the model to leverage an extended memory of possibly all past frames, offering comprehensive motion information beyond just the most recent 2-4 generated frames. By drawing on this richer context, the model mitigates temporal error accumulation from recent artifacts, thus enhancing temporal coherence and maintaining identity consistency in talking video generation.

\subsection{Emotion-Aware Audio Module}\label{sec:audiomodule}

\begin{figure}[t]
\centering
\includegraphics[width=0.34\textwidth]{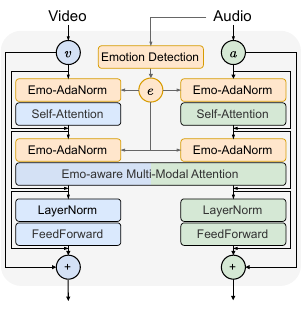}
\vspace{-0.1in}
\caption{Emotion-aware audio module.}\label{fig:audio_module}
\vspace{-0.2in}
\end{figure}

Most existing diffusion-based approaches~\cite{tian2024emo,xu2024hallo,chen2024echomimic} use cross-attention mechanisms to incorporate audio guidance in video generation, while others~\cite{xu2024vasa,tan2024flowvqtalker} apply a single, human-defined emotion label to generate emotionally talking videos during inference. However, cross attention relies on fixed audio features, limiting the depth of audio-video interaction during the diffusion process. Meanwhile, using a fixed, human-defined emotion label for the whole video fails to capture the dynamic emotional shifts in audio, resulting in facial expressions that do not naturally align with the audio emotion. To address these issues, we develop a new emotion-aware audio module to improve audio-lip consistency and align facial expressions with the audio emotion. As shown in Figure~\ref{fig:audio_module}, there are three key strategies: multi-modal attention, audio emotion-aware diffusion, and emotion decoupling training.

\paragraph{Multi-modal attention.} Our emotion-aware audio module replaces the traditional cross attention with a more dynamic multi-modal attention mechanism. Specifically, cross attention aligns video and audio by conditioning the process of video features $v$ on audio features $a$. This approach can be formalized as minimizing the loss function $\mathcal{L}_{\theta_{v|a}} = \mathbb{E}_{t,\epsilon \sim \mathcal{N}(0, I)} [ \lambda(t) \| \epsilon_\theta(v_t | a) - \epsilon \|_2^2 ]$.  In contrast,  as shown in Figure~\ref{fig:audio_module}, we explore multi-modal attention, which jointly processes both video and audio inputs by minimizing the loss function $\mathcal{L}_{\theta_{va}} = \mathbb{E}_{t,\epsilon \sim \mathcal{N}(0, I)} [ \lambda(t) \| \epsilon_\theta(v_t, a) - \epsilon \|_2^2 ]$. Such a mechanism enables better video-audio interaction during the diffusion process. 

\paragraph{Audio emotion-aware diffusion.} We then dynamically detect audio emotions to guide audio-video interaction via a newly trained emotion detection model. Specifically, the model is trained on a diverse dataset to extract emotion  $e$ from audio (See Appendix~\ref{appendix:emotion_detection} for more details), recognizing eight distinct emotions: \verb|angry|, \verb|disgusted|, \verb|fearful|, \verb|happy|, \verb|neutral|, \verb|sad|, \verb|surprised|, and \verb|others|. To improve the robustness of the detected emotion labels, emotion detection is conducted at the audio subsegment level. Each subsegment’s emotion is determined by the most frequently detected emotion across all its frames, where each frame’s emotion is evaluated using audio features from a 3-second sliding window centered on that frame. 

The detected emotion of each subsegment is then projected into emotion embeddings, and integrated into each layer via emotion-adaptive layer normalization (cf. Figure~\ref{fig:audio_module}) to guide multi-modal attention. This process results in the following emotion-conditioned flow loss:
\begin{equation}\label{eq:emo-flow}
\mathcal{L}_{\theta_{va|e}} = \mathbb{E}_{t,\epsilon \sim \mathcal{N}(0, I)} [ \lambda(t) \| \epsilon_\theta(v_t, a | e) - \epsilon \|_2^2 ].
\end{equation}

During inference, we use classifier-free guidance \cite{ho2022classifier} to control the impact of the dynamically detected emotion on the generated output. The emotion-aware output is
\begin{equation}
\tilde{\epsilon}_{\theta}( v_t, a | e) = (1 + w) \epsilon_{\theta}(v_t,a | e) - w \epsilon_{\theta}( v_t,a),
\end{equation}
where $w$ is the classifier-free guidance scale to control the influence of the emotion condition. Please note that the overall emotional tone of the generated talking video is largely inferred from the facial expression of the reference image. Here, the audio emotion $e$ aims to function mainly as a subtle adjustment to enhance or moderately alter the emotion when prompted by the audio. It is not intended to induce a complete emotional shift that would override the facial expression conveyed by the reference image.

\paragraph{Emotion decoupling training.}  To further improve the effect of audio emotion on talking videos, we introduce an emotion decoupling training strategy that separates the expression in the reference image from the audio emotion. Specifically, for training video clips sourced from MEAD~\cite{wang2020mead}—which provides both speaker identity and emotion labels—we avoid using a reference image from the same video clip. Instead, we randomly select a reference image of the same person but with a different emotion. This encourages a better disentanglement between the reference image’s expression and the audio-induced emotion, allowing our emotion-aware audio module to better refine facial expressions in alignment with the audio. Moreover, our method also supports replacing the detected audio emotion label with a manually specified emotion label, if desired.

\subsection{Data Processing Pipeline}\label{sec:datapipeline}
 
We collect a comprehensive set of open-source datasets, such as HDTF~\cite{zhang2021flow}, VFHQ~\cite{xie2022vfhq}, CelebV-HQ~\cite{zhu2022celebv}, MultiTalk~\cite{sung2024multitalk}, and MEAD~\cite{wang2020mead}, along with additional data collected by ourselves. The total duration of these raw videos exceeds 2,200 hours. However, as illustrated in Appendix~\ref{app:datapipeline}, we find that the overall quality of the data is poor, with numerous issues such as audio-lip misalignment, missing heads, multiple heads, occluded faces by subtitles, extremely small face regions, and low resolution. Directly using these data for model training results in unstable training, poor convergence, and terrible generation quality.

To further obtain high-quality talking head data, we developed a dedicated data processing pipeline for talking head generation. The pipeline consists of five steps: First, we perform scene transition detection and trim video clips to a length of less than 30 seconds. Second, we apply face detection, filtering out videos with no faces, partial faces, or multiple heads, and use the resulting bounding boxes to extract talking heads. Third, we use an Image Quality Assessment model~\cite{Su_2020_CVPR} to filter out low-quality and low-resolution videos. Fourth, we apply SyncNet~\cite{prajwal2020lip} to remove videos with audio-lip synchronization issues. Lastly, we manually assess the audio-lip synchronization and overall video quality for a subset of the data to ensure more accurate filtering. After completing the entire pipeline, the total duration of our processed high-quality videos is about 660 hours.  

\subsection{Training Strategy Decomposition}\label{sec:trainingstrategy}
  
The training of \name is divided into two progressive stages, each with specific objectives.

\noindent \textbf{Stage 1: Face domain adaptation.} Following~\cite{tian2024emo,xu2024hallo,chen2024echomimic}, we initialize Reference Net and the spatial module of Diffusion Net with the weights of SD 1.5~\cite{rombach2022high}. In this stage, we adapt Reference Net, the spatial attention modules of Diffusion Net, and the original text cross-attention module to the face domain with the rectified flow loss (cf. Eq.~\ref{eqn:flow}), ensuring these components capture facial features effectively.

\noindent \textbf{Stage 2: Emotion-decoupled robust training.} We then integrate the emotion-aware audio module and memory-guided temporal module into the Diffusion Net. Initially, we perform a warm-up training phase for the newly added modules, keeping the modules in Stage 1 fixed. After the warm-up, all modules are jointly trained. In this stage, we use the emotion-conditioned flow loss (cf. Eq.~\ref{eq:emo-flow}) and scale up the dataset to include all processed data for more comprehensive training. Here, we adopt the emotion decoupling training strategy (cf. Section~\ref{sec:audiomodule}) only when the training video clips are sourced 
from MEAD. Moreover, we found that some noisy data persisted even after applying our data processing pipeline (cf.~Section~\ref{sec:datapipeline}), making diffusion training unstable and leading to biased model optimization. To mitigate this, we further develop a robust training strategy that filters out data points with loss values suddenly exceeding a specific large value (0.1 in our case), as the emotion-conditioned flow loss in our method typically converges and fluctuates around 0.03.

\section{Experiments}

\begin{table}[t]
    \centering
    \caption{Quantitative results of video quality and audio-lip synchronization on two OOD test datasets. \name consistently outperforms existing talking video baselines.} 
\vspace{-0.1in} 
 \begin{threeparttable} 
    \centering 
     \scalebox{0.7}{ 
    \begin{tabular}{l|ccc|ccc}
    \toprule 
    \multirow{2}{*}{\textbf{Method}} & \multicolumn{3}{c|}{\textbf{VoxCeleb2 test set}} & \multicolumn{3}{c}{\textbf{Collected OOD dataset}} \cr \cmidrule{2-4} \cmidrule{5-7}
     & \textbf{FVD$\downarrow$} & \textbf{FID$\downarrow$} & \textbf{Sync-D$\downarrow$} & \textbf{FVD$\downarrow$} & \textbf{FID$\downarrow$} & \textbf{Sync-D$\downarrow$} \\ 

    \midrule 
    SadTalker~\cite{zhang2023sadtalker} & 397.0 & 71.7 & 8.6 & 288.7 & 48.3 & 10.6\\
    AniPortrait~\cite{wei2024aniportrait} & 333.2 & 45.5 & 11.0 & 238.7 & 31.2 & 10.5 \\
    V-Express~\cite{wang2024v} & 418.9 & 58.9 & 8.2 & 315.2 & 46.7 & 9.5 \\
    Hallo~\cite{xu2024hallo} & 330.4 & 41.6 & 8.0 & 231.1 & 31.9 & 9.3 \\
    Hallo2~\cite{cui2024hallo2} & 302.0 & 41.6 & 8.0 & 223.1 & 29.8 & 9.3 \\
    EchoMimic~\cite{chen2024echomimic} & 293.9 & 43.8 & 10.1 & 223.9 & 39.9 & 9.8 \\
      \midrule 
    \textbf{\name ({Ours})} & \textbf{254.3} & \textbf{31.7} & \textbf{7.4} & \textbf{161.1} & \textbf{24.9} & \textbf{9.2} \\
    
    \bottomrule 
    \end{tabular}}
	 \end{threeparttable}
    \label{tab:combined_distribution}
    \vspace{-0.1in}
\end{table}

\subsection{Experimental Setup}

\noindent \textbf{Evaluation benchmarks.} We create two out-of-distribution (OOD) datasets to evaluate \name's performance and generalization capabilities. For the first OOD dataset, We sample 150 video clips from the VoxCeleb2~\cite{nagrani2020voxceleb} test set, which contains videos of various celebrities. Similarly, we create the second OOD dataset with 150 clips across a more diverse set of audio, backgrounds, ages, genders, and languages.

\noindent \textbf{Evaluation metrics.} We adopt a suite of metrics to evaluate the overall quality and audio-lip synchronization of the generated videos. The Fréchet Video Distance (FVD)~\cite{unterthiner2019fvd} measures the distance between the distributions of real and generated videos, providing an assessment of overall video quality. The Fréchet Inception Distance (FID)~\cite{heusel2017gans} evaluates the quality of individual frames by comparing feature distributions extracted from a pre-trained model. SyncNet Distance (Sync-D)~\cite{chung2017out} measures audio-lip synchronization using a pre-trained discriminator model.  

\noindent \textbf{Baselines.} We compare our method against several state-of-the-art baselines with publicly available model checkpoints. The baselines include both two-stage methods with intermediate representations and end-to-end diffusion methods. V-Express~\cite{wang2024v},  AniPortrait~\cite{wei2024aniportrait} and EchoMimic \cite{chen2024echomimic} are two-stage methods using intermediate representations like landmarks, while Hallo~\cite{xu2024hallo} and Hallo2~\cite{cui2024hallo2} are recent end-to-end diffusion-based models. More implementation details of \name are in Appendix~\ref{app:implementation}.

\begin{figure}
    \centering
\includegraphics[width=0.47\textwidth]{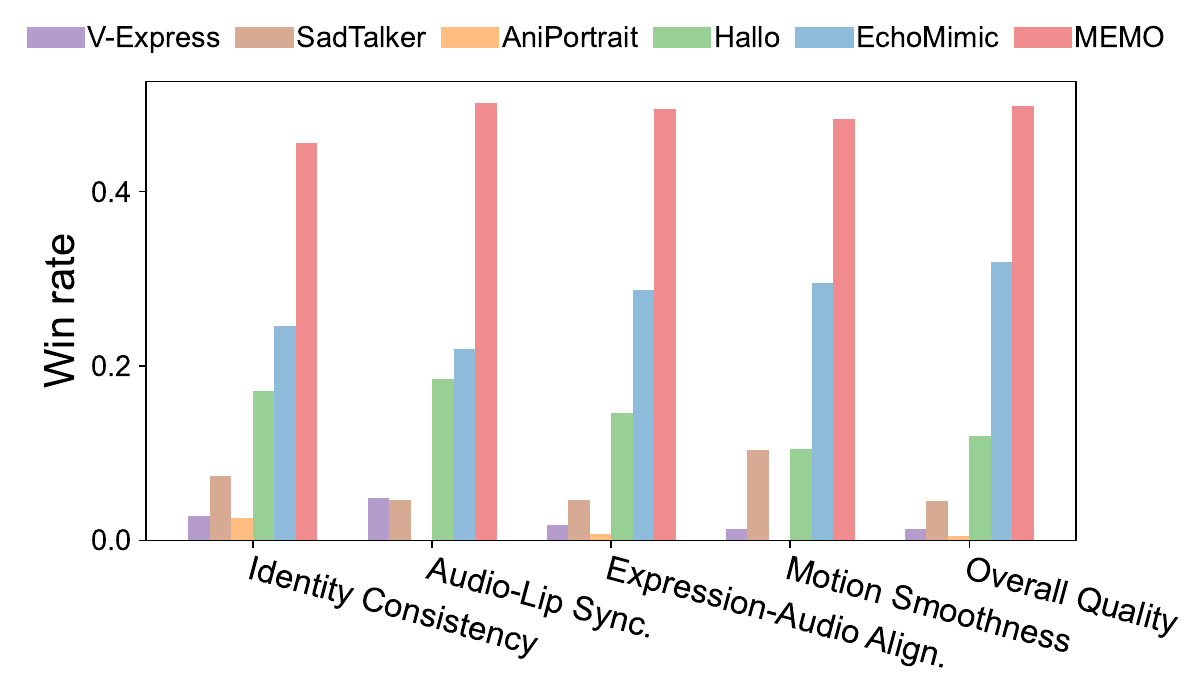} 
 \vspace{-0.2in}
\caption{Human preferences among \name and baselines, where users select the best method in terms of each evaluation metric.}
\label{fig:human_study}
    \vspace{-0.2in}
\end{figure}

\begin{figure*}[t]
    \vspace{-0.2in}
    \centering
    \includegraphics[width=0.85\textwidth]{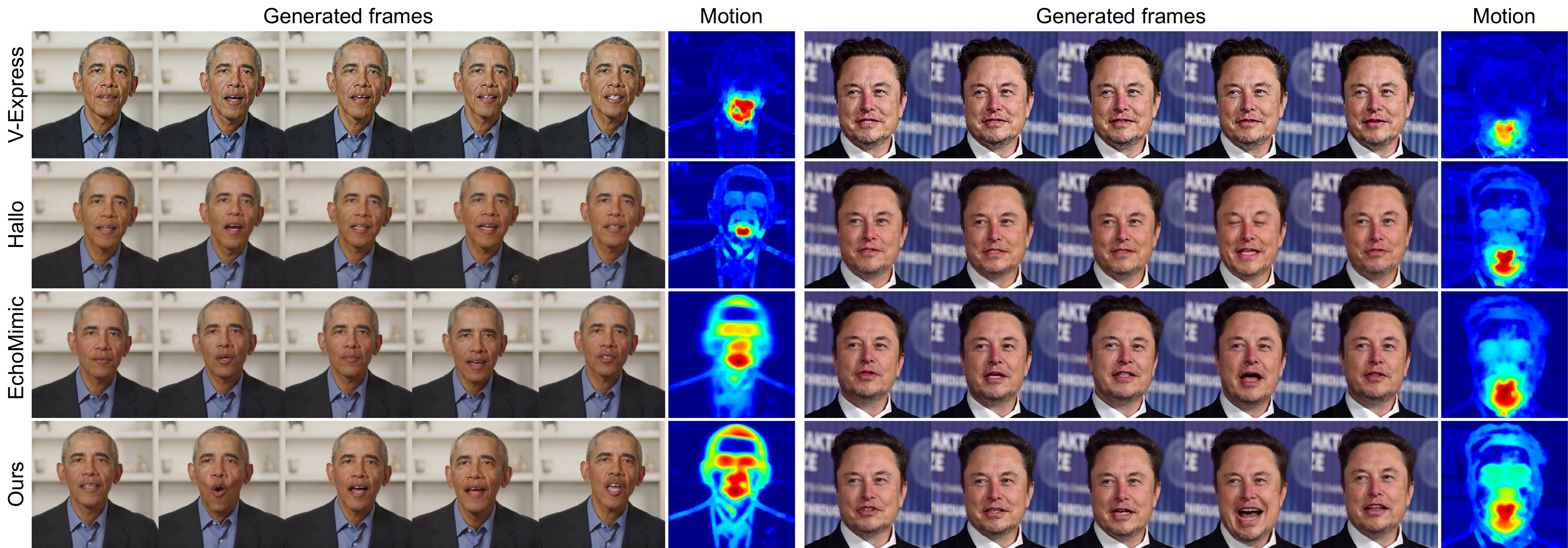}
    \caption{\name can generate talking videos featuring a wider range of smooth head movements and more emotional facial expressions, illustrated in both visualization and heatmaps. Please refer to the supplementary material for video demos.}
    \label{fig:visualization}
    \vspace{-0.1in}
\end{figure*}

\begin{figure*}[t]
  \centering
  \begin{minipage}{0.48\textwidth}
    \centering
    \includegraphics[width=0.9\textwidth]{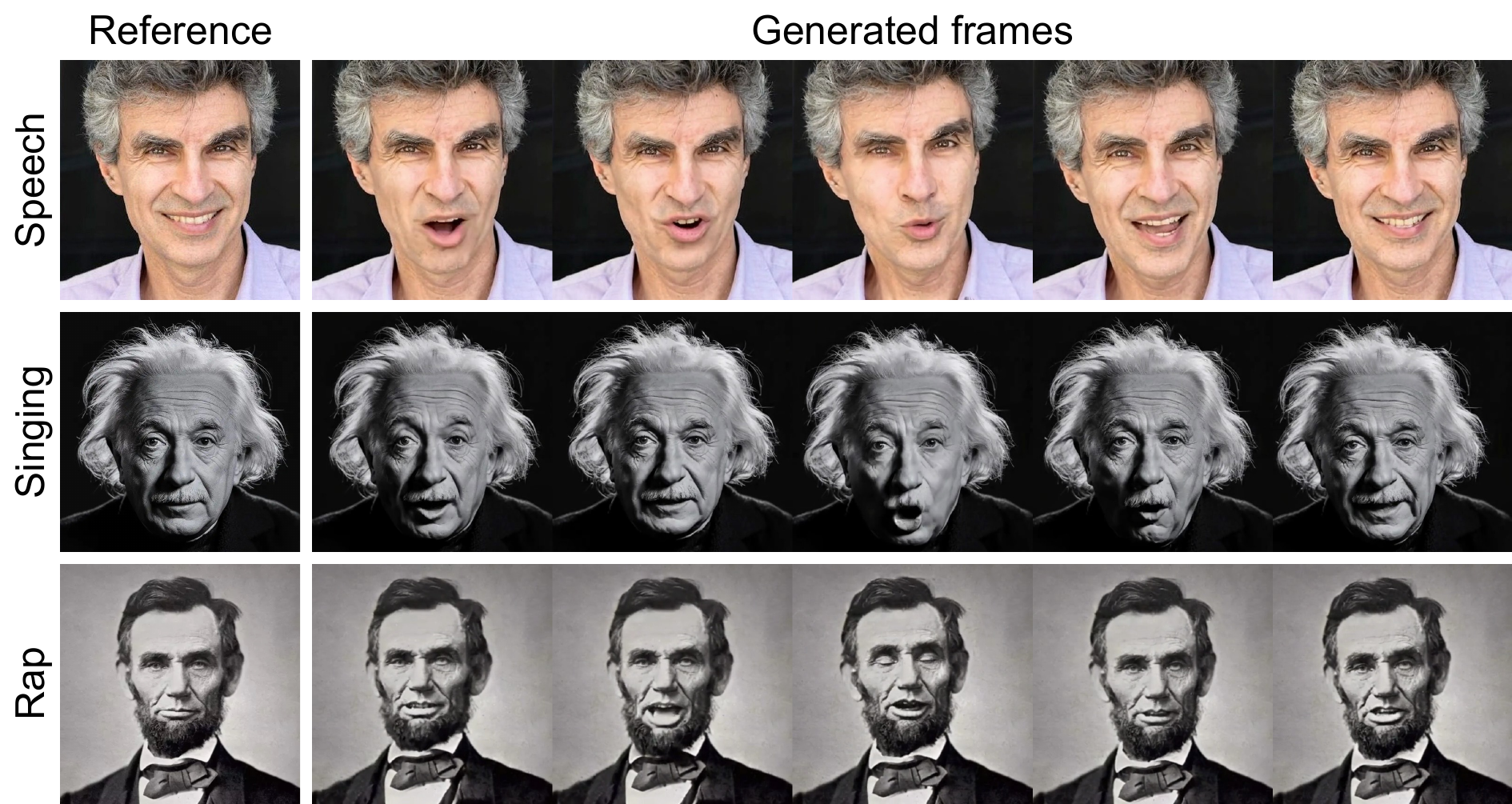}
    \caption{The generated videos with various types of driving audio. Please refer to the supplementary material for video demos.}
    \label{fig:special_audio_demo}
  \end{minipage}%
  \hfill
  \begin{minipage}{0.48\textwidth}
    \centering
    \includegraphics[width=0.9\textwidth]{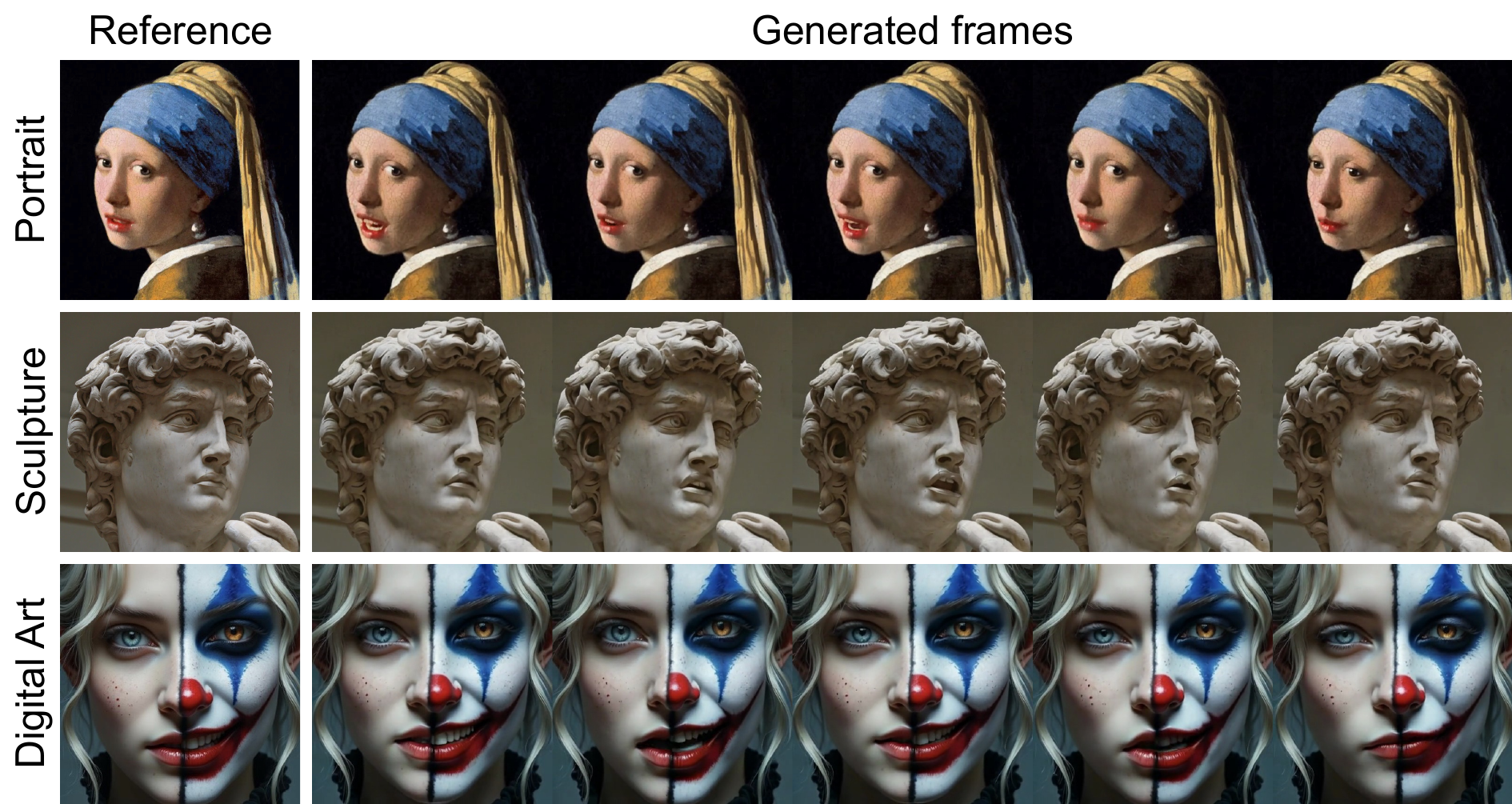}
    \caption{The generated videos with various types of reference images. Please refer to the supplementary for video demos.}
    \label{fig:special_refimg_demo}
  \end{minipage} 
\end{figure*}

\subsection{Quantitative Results}
\label{sec:quant}

\noindent \textbf{Performance on two OOD test sets.} Table \ref{tab:combined_distribution} reports the quantitative results on two OOD test sets. Our method consistently outperforms all baselines in terms of FVD, FID, and Sync-D metrics, indicating better video quality and audio-lip synchronization. These results also demonstrate the improved generalization abilities of \name to unseen identities and audio. 

\noindent \textbf{Human evaluation.} To better benchmark the quality of generated talking videos, we conduct human studies based on five subjective metrics in several challenging scenarios, \emph{e.g.}, singing, rap, and multi-lingual talking video generation. Specifically, our analyses are based on the overall quality, motion smoothness, expression-audio alignment, audio-lip synchronization, and identity consistency. As shown in Figure~\ref{fig:human_study}, our method achieves the highest scores across all criteria in human evaluations, much higher than compared methods. This further demonstrates the effectiveness of our approach.

\subsection{Qualitative Results}

\noindent \textbf{Diversity of head motions.} Figure~\ref{fig:visualization} shows that \name can generate talking videos with higher diversity in head motions, compared to existing methods. The improved motion diversity contributes to better naturalness and expressiveness of talking videos.

\noindent \textbf{Generalization to different types of audio.} We evaluate \name on various audio types, including speeches, songs, and raps. In Figure~\ref{fig:special_audio_demo}, \name consistently generates synchronized lip movements across diverse audio types. It performs well with both expressive songs, which demand nuanced emotional alignment, and raps, which require rapid audio-lip synchronization. This verifies \name's strong generalization to various types of driving audio. 

\noindent \textbf{Generalization to different styles of reference images.} Figure~\ref{fig:special_refimg_demo} shows \name's performance on challenging reference images of diverse styles, such as portraits, sculpture, and digital art images. Despite these styles deviating significantly from our training data, \name maintains robust generation quality without producing noticeable artifacts, demonstrating its ability to generalize to various OOD reference images.

\begin{figure*}[t]
  \vspace{-0.2in}
  \centering
  \begin{minipage}{0.45\textwidth}
    \centering
    \includegraphics[width=0.87\textwidth]{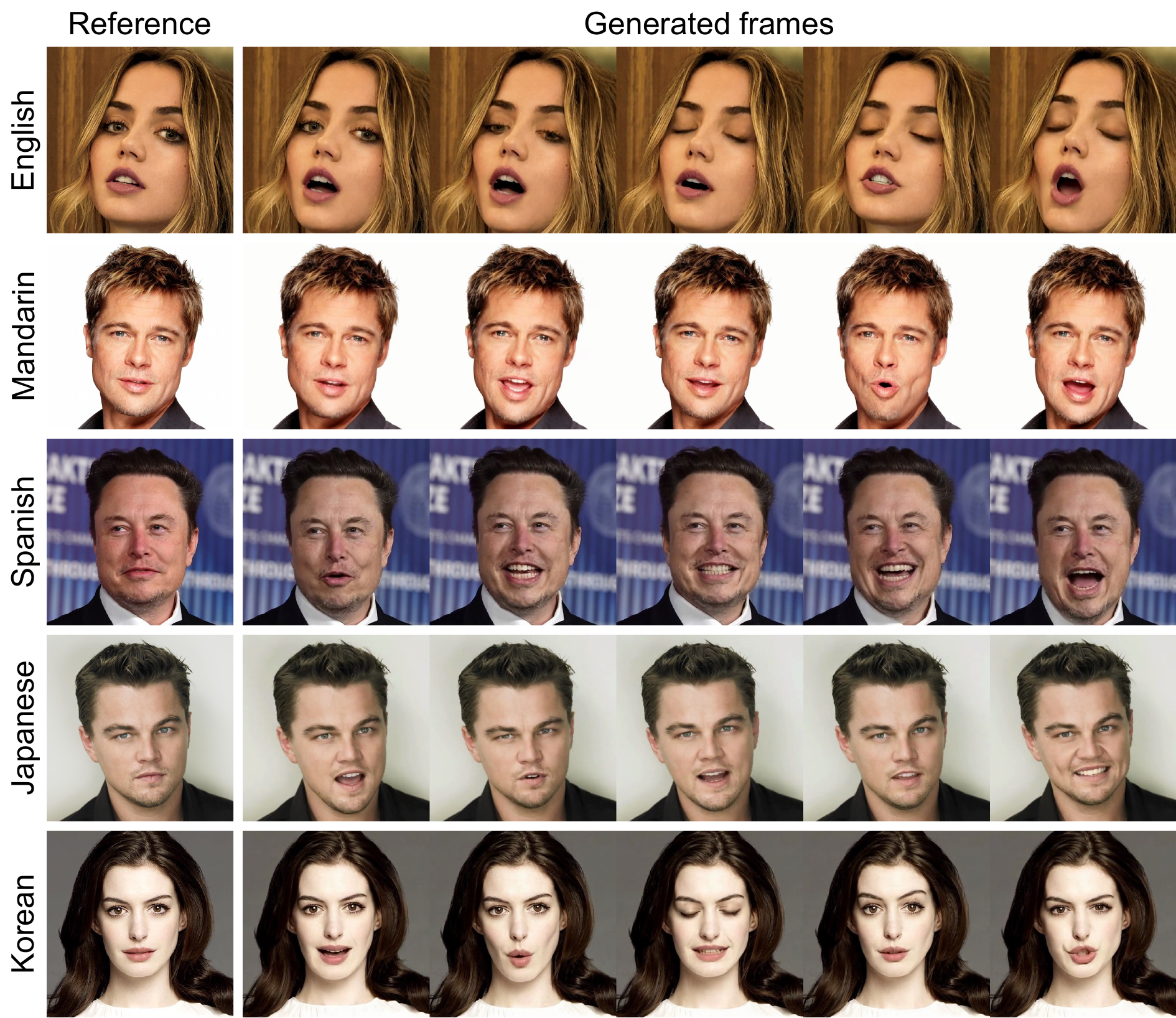}
\vspace{-0.1in}
    \caption{The generated videos on driving audio with different languages. See the supplementary for video demos.}
    \label{fig:multilingual_demo}
  \end{minipage}%
  \hfill
  \begin{minipage}{0.45\textwidth}
    \centering
    \includegraphics[width=0.87\textwidth]{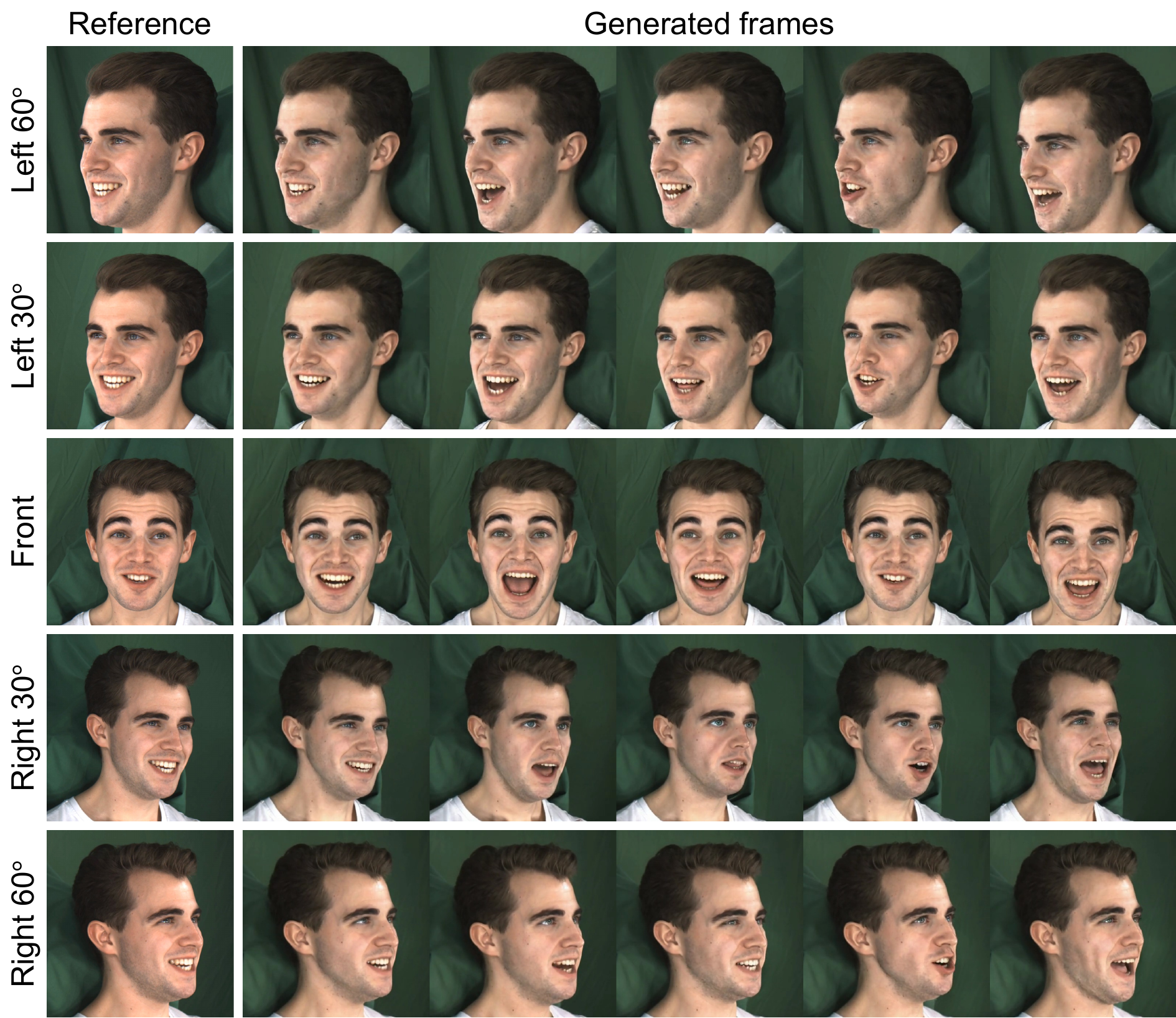}
\vspace{-0.1in}
    \caption{The generated videos with reference images of different head poses. See the supplementary for video demos.}
    \label{fig:sideface_demo}
  \end{minipage}
  \vspace{-0.1in}
\end{figure*}

\begin{figure*}[t]
  %\vspace{-0.1in}
  \centering
  \begin{minipage}{0.45\textwidth}
    \centering
    \includegraphics[width=\textwidth]{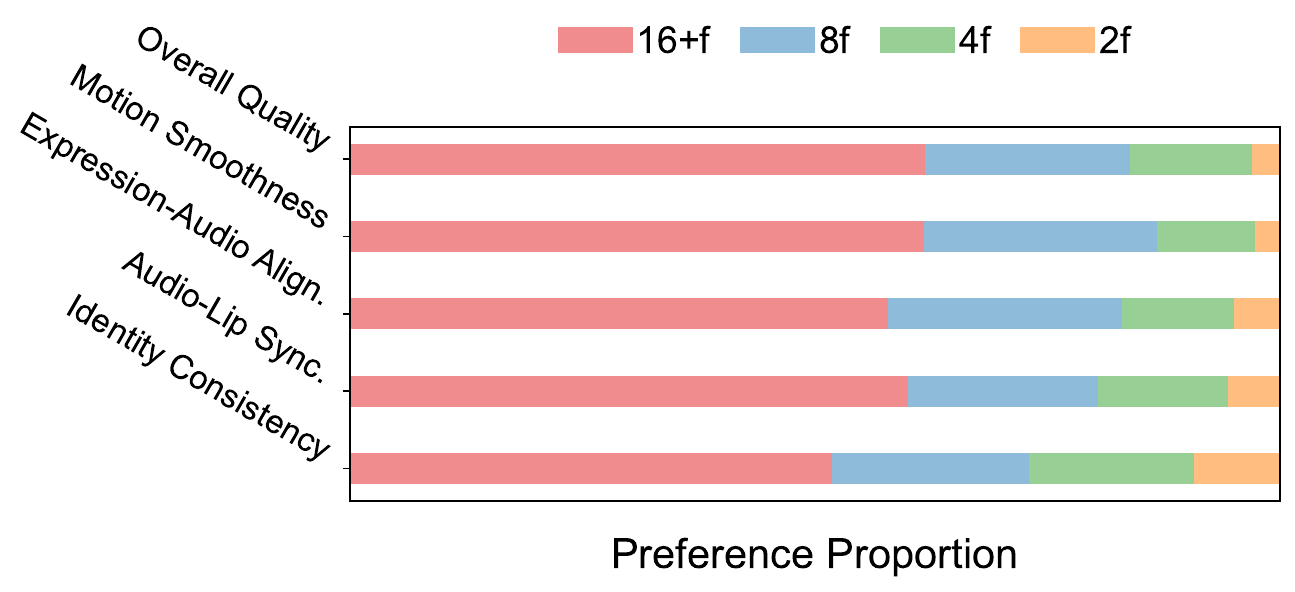}
    \vspace{-0.3in}
    \caption{Ablation on the number of past frames ($f$) during inference via human evaluation, where 16+f indicates our memory-guided inference with a context beyond 16 frames.}
    \label{fig:ablate_f_human}
  \end{minipage}%
  \hfill
  \begin{minipage}{0.45\textwidth}
    \centering
    \includegraphics[width=\textwidth]{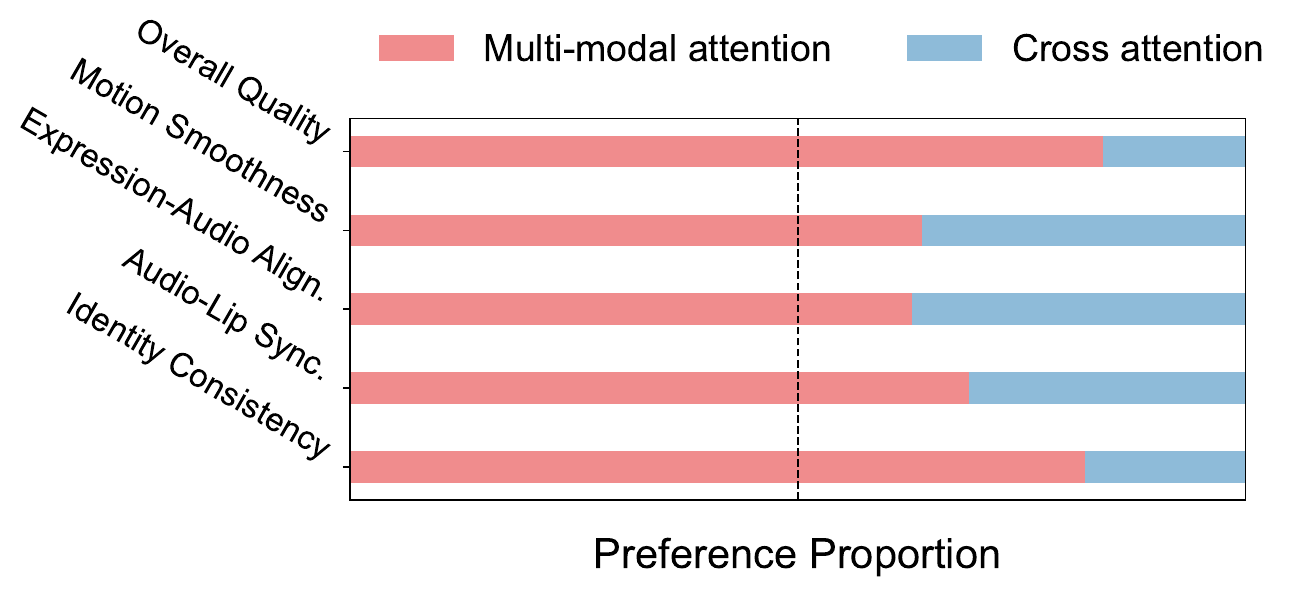}
    \vspace{-0.3in}
    \caption{Human preference comparison between multi-modal attention and cross attention for integrating audio conditions in the audio module.}
    \label{fig:mm_dif_human}
  \end{minipage}
  \vspace{-0.1in}
\end{figure*}

\noindent \textbf{Generalization to multilingual audio.} As shown in Figure~\ref{fig:multilingual_demo}, our method demonstrates robust generalization across multilingual audio inputs, such as English, Chinese, Spanish, Japanese, and Korean. Despite most of our training data being in English, our approach effectively generates lip movements synchronized with the given multilingual audio while capturing rich, realistic facial expressions.

\noindent \textbf{Generalization to different head poses.} Figure~\ref{fig:sideface_demo} shows the generated videos of \name with reference images of varying head poses, including frontal views and multiple side angles. This demonstrates that our method can generate realistic talking videos across different angles while maintaining consistency in facial appearance and expression. 

\subsection{Ablation Studies}\label{sec_ablation}

To more comprehensively evaluate the effects of our method's components, we conduct ablation studies based on human evaluation and qualitative analysis.

\noindent \textbf{Effects of memory module.} We evaluate the impacts of our memory module by ablating the length of past frames as temporal guidance. As shown in Figure~\ref{fig:ablate_f_human}, longer memory significantly improves temporal coherence, overall quality, motion smoothness, identity consistency, and audio-lip alignment, while short motion frames lead to worse performance. This demonstrates the effectiveness of our memory-guided temporal module and also explains why our method can alleviate temporal error accumulation in Figure~\ref{fig:demo}, whereas Hallo2 is prone to error accumulation.

\noindent \textbf{Effects of multi-modal attention.} We further investigate the impact of the multi-modal attention through human evaluations. Results in Figure~\ref{fig:mm_dif_human} underscore the effectiveness of multi-modal attention over cross attention in terms of the overall video quality and audio-lip alignments.

\begin{figure}[t]
 \vspace{-0.1in}
    \centering
    \includegraphics[width=0.4\textwidth]{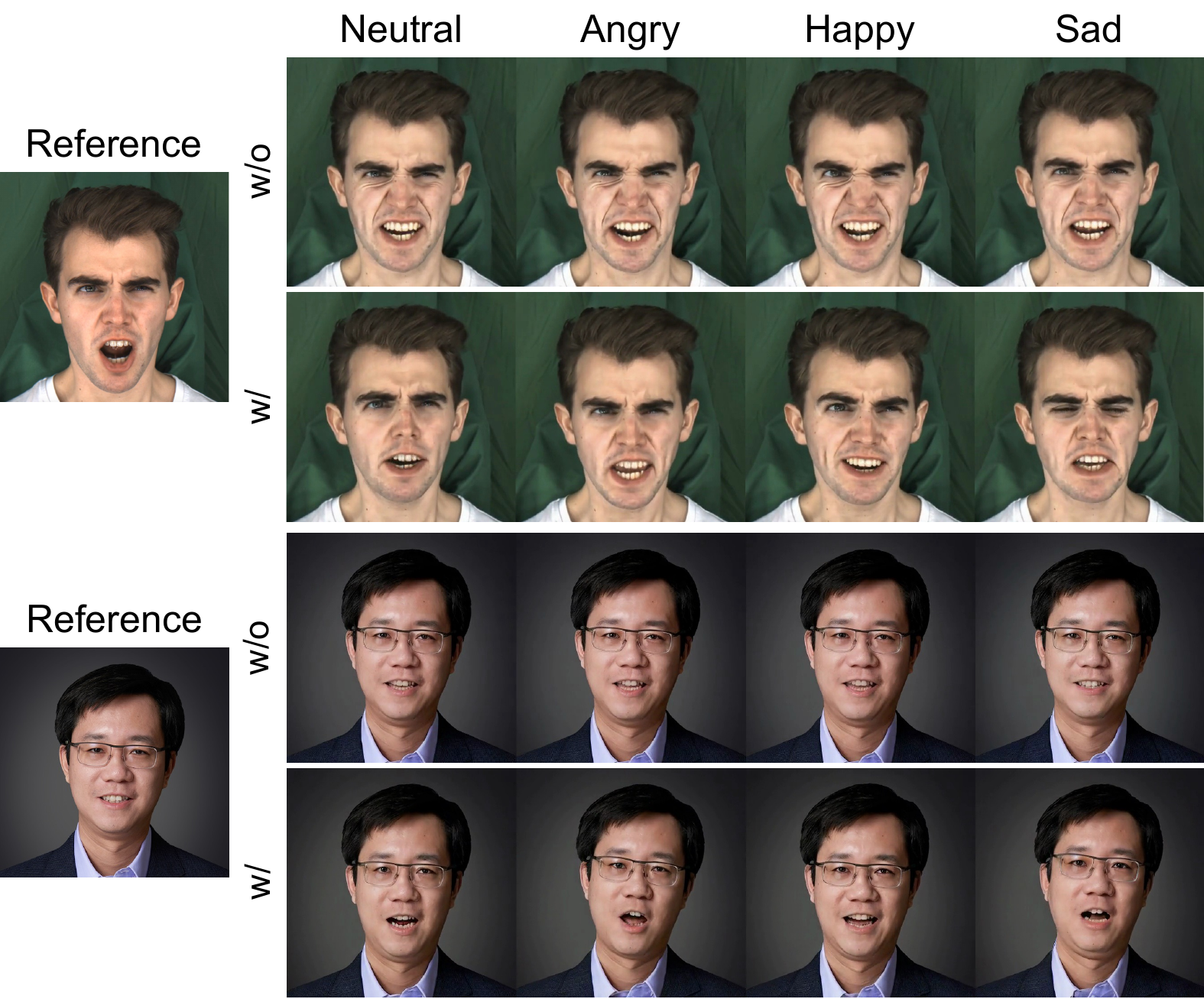}
\vspace{-0.1in}
    \caption{Ablation of our emotion-aware module with or without emotion decoupled training, given various human-defined emotion labels. Refer to the supplementary for video demos.}
    \label{fig:emotion_demo}
\vspace{-0.2in}
\end{figure}

\noindent \textbf{Effects of emotion-aware scheme.} To investigate the effects of our emotion-aware scheme (\ie emotion-aware diffusion and emotion decoupling training), we conduct an ablation study by manually replacing the detected audio emotion with a fixed, human-defined emotion label. In Figure~\ref{fig:emotion_demo}, using the same reference image and different emotion labels, we compare the frames generated by our method with and without emotion decoupling training at the same audio moment. The results demonstrate that our emotion-aware module with emotion decoupling training can effectively refine facial expressions to align with the specified emotion labels. This finding suggests that, with our detected emotion labels, \name can generate facial expressions that match the audio emotion. Moreover, this comparison also validates the necessity of our emotion decoupling training.

\section{Conclusion}

This work has presented \name, a state-of-the-art talking video generation method. Specifically,  \name effectively alleviates temporal error accumulation and enhances long-term identity consistency  via a new memory-guided temporal module, while generating videos with high audio-lip and expression-audio alignment via a new emotion-aware audio module.  Moreover, \name does not need face-related inductive biases in the model architecture, allowing it to be extended to broader applications, such as talking body generation tasks. In the future, it is interesting to explore Diffusion Transformer~\cite{peebles2023scalable} with better identity preservation strategies for talking video generation.

{
\clearpage
\small
\bibliographystyle{ieeenat_fullname}
\bibliography{main}
}

\clearpage
\input{appendix}


\end{document}

%% file: preamble.tex
\usepackage[T1]{fontenc}

\usepackage{subcaption}
\usepackage{url}
\usepackage{graphicx}
\usepackage{xspace}
\usepackage{booktabs}
\usepackage{wrapfig} 
\definecolor{cvprblue}{rgb}{0.21,0.49,0.74}
\usepackage[pagebackref,breaklinks,colorlinks,allcolors=cvprblue]{hyperref}
\hypersetup{
 colorlinks=true,
 linkcolor=red,
 citecolor=cyan,
 filecolor=magenta, 
 urlcolor=cyan,
}
\usepackage{multicol,multirow} 
\usepackage{threeparttable}  

\newcommand{\name}[0]{\mbox{MEMO}\xspace}

%% file: appendix.tex
\appendix
%\setcounter{page}{1}
% \maketitlesupplementary

\section{More Related Studies on Diffusion Models}\label{app:related}

Diffusion models~\citep{sohl2015deep,ho2020denoising} are highly expressive generative models, demonstrating remarkable capabilities in image synthesis~\citep{rombach2022high,podell2023sdxl} and video generation~\citep{guo2023animatediff,xing2023dynamicrafter}. Stable Diffusion~\cite{rombach2022high} employs a UNet architecture and generates high-resolution images in the latent space, which is extended to video domains by AnimateDiff~\citep{guo2023animatediff} by adding temporal attention layers. These models generate images or videos based on text prompts, where the text guidance from the pre-trained text encoder is introduced through cross attention. In the domain of talking video generation, diffusion models also show promising results in generation quality~\citep{he2023gaia,tian2024emo,wei2024aniportrait,xu2024hallo,stypulkowski2024diffused,xu2024vasa}, outperforming previous GAN-based methods~\citep{prajwal2020lip,zhou2020makelttalk}. Instead of using text prompts, most of these diffusion-based methods condition diffusion models on image and audio embeddings extracted from a pre-trained image encoder and audio encoder, respectively.

\section{More Implementation Details}\label{app:implementation}

Both Reference Net and the spatial module of Diffusion Net are initialized with the weights of SD 1.5~\citep{rombach2022high}.  The temporal module of Diffusion Net is initialized from AnimateDiff~\citep{guo2023animatediff}. Here, the Reference Net provides identity information from reference images for spatial modeling of the Diffusion Net and offers temporal information from past frames for temporal modeling of the Diffusion Net. For both Reference Net and Diffusion Net, we replace the text cross-attention with image cross-attention. We add two projection modules to convert the audio embedding and image embedding into the dimensions required by our attention module. The audio embedding consists of all the hidden states from the Wav2Vec 2.0 model~\citep{baevski2020wav2vec}.

The training videos are center-cropped and resized to a resolution of 512 $\times$ 512 pixels. With a fixed learning rate of 1e-5, we train \name for 15k and 600k steps at training stages 1 and 2, respectively. 
During Stage 2, a fixed number of 16 past frames is used to compute memory states as motion context. Alternatively, the number of past frames can be dynamically chosen from 16, 32, or 48 for training, enabling the model to handle longer context scenarios more effectively.  Nevertheless, thanks to the memory update mechanism with causal history decay (cf. Section~\ref{sec:temporalmodule}), such dynamic past-frame training is unnecessary. Moreover, emotion embeddings, reference images, audio embeddings, and past frames are randomly dropped with a probability of 5\% for classifier-free inference.  At inference, we set the frame rate to 30 frames per second (FPS) and employ autoregressive generation, producing 16 frames per iteration. The classifier-free guidance scale is set to 3.5.

\section{Audio Emotion Detection}
\label{appendix:emotion_detection}

To improve the natural expression of talking videos, we develop an emotion detection model to detect emotion labels from audio. In this appendix, we first introduce the data collection and processing strategies for audio emotion recognition, followed by the details of our emotion detector.

\begin{table}[t]
\centering
\caption{Statistics of the emotion detection Dataset. The source column represents the origin of the samples, and the language column specifies the dataset's language. \#Emo indicates the number of emotion categories, \#Utts shows the total number of utterances, and \#Hrs represents the total hours of training data}
\label{tab:emotion-dataset}
\scalebox{0.7}{  
\begin{tabular}{l l l r r r}
\toprule
\multicolumn{6}{c}{\textbf{Speech Emotion Recognition datasets}} \\
\midrule
\textbf{Dataset} & \textbf{Source} & \textbf{Language} & \textbf{\#Emo} & \textbf{\#Utts} & \textbf{\#Hrs} \\
\midrule
AESDD~\citep{aesdd}        & Act     & Greek    & 5   & 604 & 0.7   \\
ASED~\citep{ased}          & Act     & Amharic  & 5   & 2,474 & 2.1   \\
ASVP-ESD~\citep{asvpesd}  & Media   & Mix      & 12  & 13,964 & 18.0  \\
CaFE~\citep{cafe}          & Act     & French   & 7   & 936  & 1.2   \\
%CASIA [10]       & Act     & Mandarin & 6  & 4   & 1200    & 0.6   \\
%CREMA-D [11]     & Act     & English  & 6  & 91  & 7442    & 5.3   \\
EMNS~\citep{emns}          & Act     & English  & 8  & 1,181  & 1.9   \\
EmoDB~\citep{emodb}        & Act     & German   & 7  & 535   & 0.4   \\
EmoV-DB~\citep{emovdb}     & Act     & English  & 5  & 6,887  & 9.5   \\
%EMOVO [15]       & Act     & Italian  & 7  & 6   & 588     & 0.5   \\
Emozionalmente~\citep{emozionalmente} & Act & Italian & 7 & 6,902 & 6.3 \\
eNTERFACE~\citep{enterface}  & Act  & English  & 6   & 1,263  & 1.1   \\
ESD~\citep{esd}   & Act     & Mix      & 5   & 35,000   & 29.1  \\
%IEMOCAP [19]     & Act     & English  & 5  & 10  & 5531    & 7.0   \\
JL-Corpus~\citep{jlcorpus}   & Act     & English  & 5  & 2,400 & 1.4  \\
M3ED~\citep{m3ed}    & TV      & Mandarin & 7 & 24,437   & 9.8   \\
MEAD~\citep{wang2020mead}   & Act     & English  & 8  & 31,729   & 37.3  \\
%MELD [23]        & TV      & English  & 7  & 304 & 13706   & 12.1  \\
%MER2023 [24]     & TV      & Mandarin & 6  & /   & 5030    & 5.9   \\
MESD~\citep{mesd}  & Act     & Spanish  & 6   & 862  & 0.2   \\
%MSP-Podcast [26] & Podcast & English  & 8  & 1273& 73042   & 113.6 \\
Oreau~\citep{oreau}  & Act   & French   & 7   & 434  & 0.3   \\
PAVOQUE~\citep{pavoque} & Act  & German   & 5   & 7,334  & 12.2  \\
Polish~\citep{polish}  & Act  & Polish   & 3    & 450     & 0.1   \\
RAVDESS~\citep{ravdess} & Act  & English  & 8  & 1,440    & 1.5   \\
%RESD [31]        & Act     & Russian  & 7  & 200 & 1396    & 2.3   \\
SAVEE~\citep{savee}   & Act     & English  & 7   & 480     & 0.5   \\
%ShEMO [33]       & Act     & Persian  & 6  & 87  & 2838    & 3.3   \\
SUBESCO~\citep{subesco} & Act   & Bangla   & 7   & 7,000    & 7.8   \\
TESS~\citep{tess}   & Act     & English  & 7  & 2,800    & 1.6   \\
TurEV-DB~\citep{turevdb} & Act  & Turkish  & 4  & 1,735    & 0.5   \\
URDU~\citep{urdu}  & Talk show & Urdu  & 4   & 400     & 0.3   \\
\midrule
% \textbf{Total}   & -       & -        & -  & 3510& 262020  & 294.4 \\
\multicolumn{6}{c}{\textbf{Music Emotion Recognition Datasets}}\\
\midrule
\textbf{Dataset} & \textbf{Source} & \textbf{Lang} & \textbf{Emo} & \textbf{\#Utts} & \textbf{\#Hrs} \\
\midrule
RAVDESS-Song~\citep{ravdess} & Act & English & 6 & 1,012 & 1.31 \\
MTG-Jamendo~\citep{mtgjamendo} & Media & Mix & 56 & 5,022 & 299.47 \\

\bottomrule
\end{tabular}} 
\end{table}

\subsection{Dataset Collection and Processing}

\paragraph{Dataset collection.} To achieve robust emotion detection across both speech and music audio sources, we collect a large-scale dataset encompassing both speech and music segments, each annotated with emotion labels. A detailed overview of the datasets used in our training process is provided in Table~\ref{tab:emotion-dataset}. For speech audio, we collect data from a recent Speech Emotion Recognition benchmark, EmoBox~\citep{emobox}, which incorporates 23 datasets from various origins, covering 12 distinct languages. Regarding music audio, we gather data from the RAVDESS-song~\citep{ravdess} and MTG-Jamendo~\citep{mtgjamendo} datasets, including songs with and without background music.

All data underwent a standardized processing protocol, converted to a monophonic format with a sampling rate of 16,000 Hz. Each utterance is uniquely annotated with an emotion label. For datasets containing lengthy samples, such as MTG-Jamendo, we divide them into shorter segments of 30 seconds to align with the typically shorter length of other datasets, assigning the same label to all segments. Each dataset was then split into training and testing sets with a ratio of 3:1.

\paragraph{Label merging.} A major challenge in integrating different datasets is aligning their label spaces, as each dataset often features distinct emotion categories. For instance, the URDU dataset~\citep{urdu} contains only four emotion labels: happy, sad, angry, and neutral. In contrast, the ASVP-ESD dataset~\citep{asvpesd} includes 12 emotion labels, covering less common emotions such as boredom and pain. For music emotion recognition datasets like MTG-Jamendo~\citep{mtgjamendo}, there are 56 mood/theme tags, not all of which correspond to emotional labels, and each sample can be assigned multiple tags. These discrepancies and overlaps in category spaces across different datasets present significant challenges for emotion detection.

To establish a generalized and streamlined label space, we designed our module to perform an 8-class classification task, selecting labels that are both commonly recognized and easily distinguishable: 
\verb|angry|, \verb|disgusted|, \verb|fearful|, \verb|happy|, \verb|neutral|, \verb|sad|, \verb|surprised|, and \verb|others|. 
We meticulously reviewed and mapped the original labels from each dataset to fit within this new label space. For instance, samples labeled as \verb|pleasure| in the ASVPESD dataset were mapped to the \verb|happy| category due to their semantic similarity. Labels that did not clearly correspond to a specific emotion were categorized under the \verb|others| label.

\subsection{Audio Emotion Detector}

\begin{table}  
 
\centering
\caption{Accuracy comparison of audio emotion detection between Emotion2vec~\citep{emotion2vec} and our learned emotion detector.}
\label{tab:emotion-detection-accuracy}
 \resizebox{0.37\textwidth}{!}{
\begin{tabular}{l r r}
\toprule
\textbf{Dataset} & \textbf{Emotion2vec}& \textbf{Ours} \\
\midrule
AESDD           & 75.84 &  78.52\\
ASED            & 86.20 &  85.23\\
ASVP-ESD        & 52.55 & 55.99\\
CaFE            & 73.30 & 100.00\\
EMNS            & 57.98 & 61.87\\
EmoDB           & 88.41 & 100.0\\
EmoV-DB         & 77.84 & 91.22\\
Emozionalmente  & 66.61 & 71.02\\
eNTERFACE       & 28.21 & 32.05\\
ESD             & 94.83 & 99.94\\
JL-Corpus       & 71.92 & 100.00\\
M3ED            & 42.59 & 41.52\\
MEAD            & 61.74 & 71.45\\
MESD            & 40.65 & 41.12\\
Oreau           & 50.96 & 42.31\\
PAVOQUE         & 85.15 & 92.74\\
Polish          & 44.89 & 100.00\\
RAVDESS         & 82.36 & 100.00\\
SAVEE           & 83.33 & 100.00\\
SUBESCO         & 78.43 & 100.00\\
TESS            & 76.29 & 95.14\\
TurEV-DB        & 47.45 & 53.47\\
URDU            & 54.00 & 56.00\\
RAVDESS-Song    & 43.58 & 100.00\\
MTG-Jamendo     & 65.30 & 74.50\\
\midrule
\textbf{Total} & \textbf{68.78} & \textbf{78.26}\\
\bottomrule
\end{tabular}}
\end{table}

We implemented an 8-way classifier for our task, drawing inspiration from state-of-the-art methods in speech and music emotion detection. Our solution builds upon Emotion2vec~\citep{emotion2vec}, a robust universal speech emotion representation model. The feature extractor employs multiple convolutional layers and Transformer blocks and is trained using a teacher-student online distillation self-supervised learning approach. The feature extractor backbone of Emotion2vec is pre-trained on a large-scale multilingual speech corpus. For our classification task, we use the fixed Emotion2vec backbone as the feature extractor and train a 5-layer MLP as the classification head.

To stabilize the training process, we apply gradient clipping, constraining the gradient updates within an $l_2$ norm of 1.0. To enhance the model's generalization ability, we incorporate a contrastive learning technique~\citep{coretuning}. The test accuracy for each dataset, as well as the overall accuracy, is reported in Table~\ref{tab:emotion-detection-accuracy}. We compare with the original Emotion2vec~\cite{emotion2vec} as the baseline, where it adopted a single linear layer after the feature extraction backbone for downstream emotion detection.

\section{Data Processing Pipeline}\label{app:datapipeline}

\begin{figure}[t]
\centering 
\includegraphics[width=0.48\textwidth]{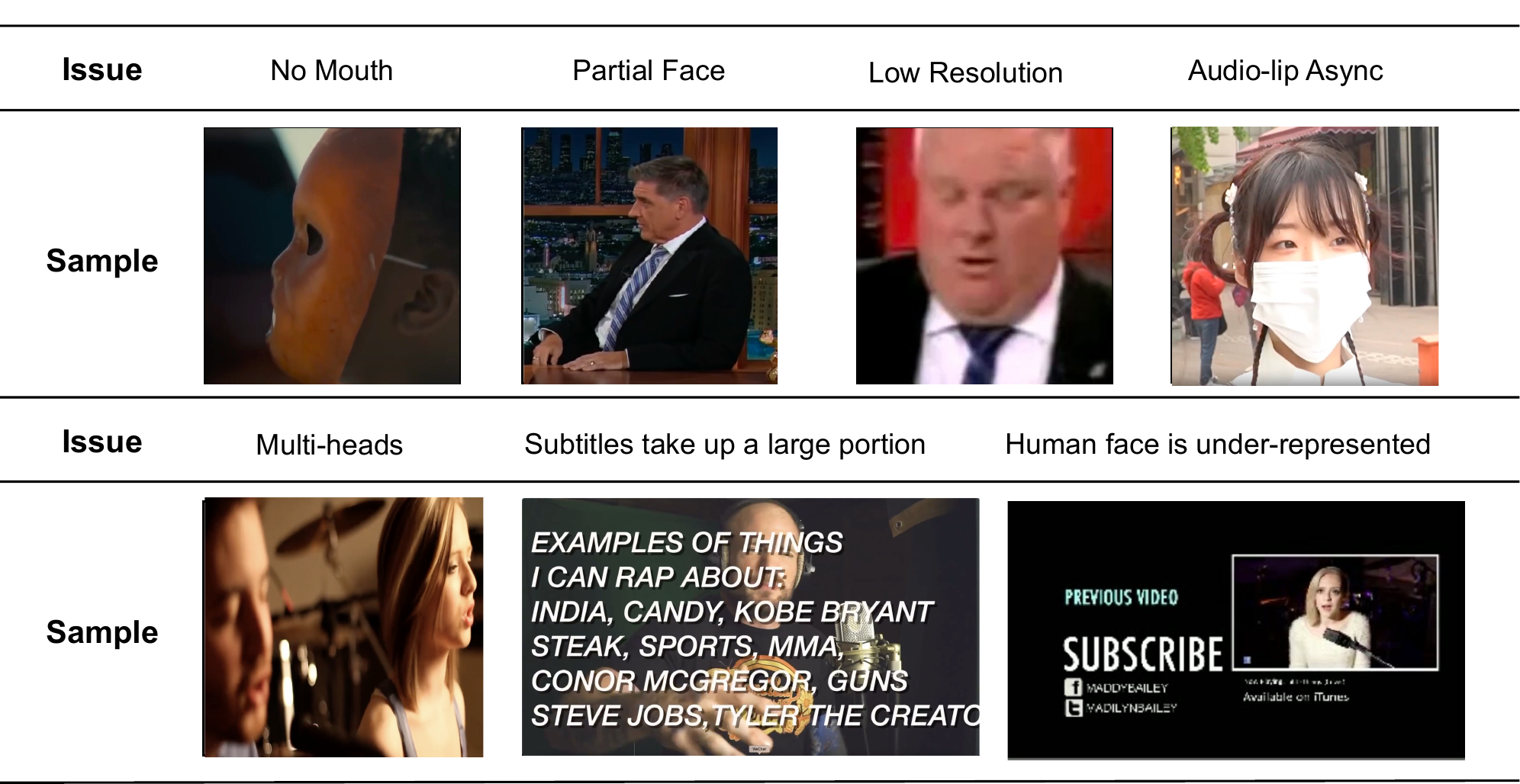}
    \vspace{-0.2in}
\caption{Examples of issues in the raw dataset.}
\label{fig:data_example}
\end{figure}

We collect a comprehensive set of open-source datasets, such as HDTF~\citep{zhang2021flow}, VFHQ~\citep{xie2022vfhq}, CelebV-HQ~\citep{zhu2022celebv}, MultiTalk~\citep{sung2024multitalk}, and MEAD~\citep{wang2020mead}, along with additional data we collected ourselves. The total duration of these raw videos exceeds 2,200 hours. However, as illustrated in Figure~\ref{fig:data_example}, we find that the overall quality of the data is poor, with numerous issues such as audio-lip misalignment, missing heads, multiple heads, occluded faces by subtitles, extremely small face regions, and low resolution. Directly using these data for model training results in unstable training, poor convergence, and terrible generation quality.

\begin{figure}[t]
    \centering
    \vspace{-0.1in}
    \includegraphics[width=0.45\textwidth]{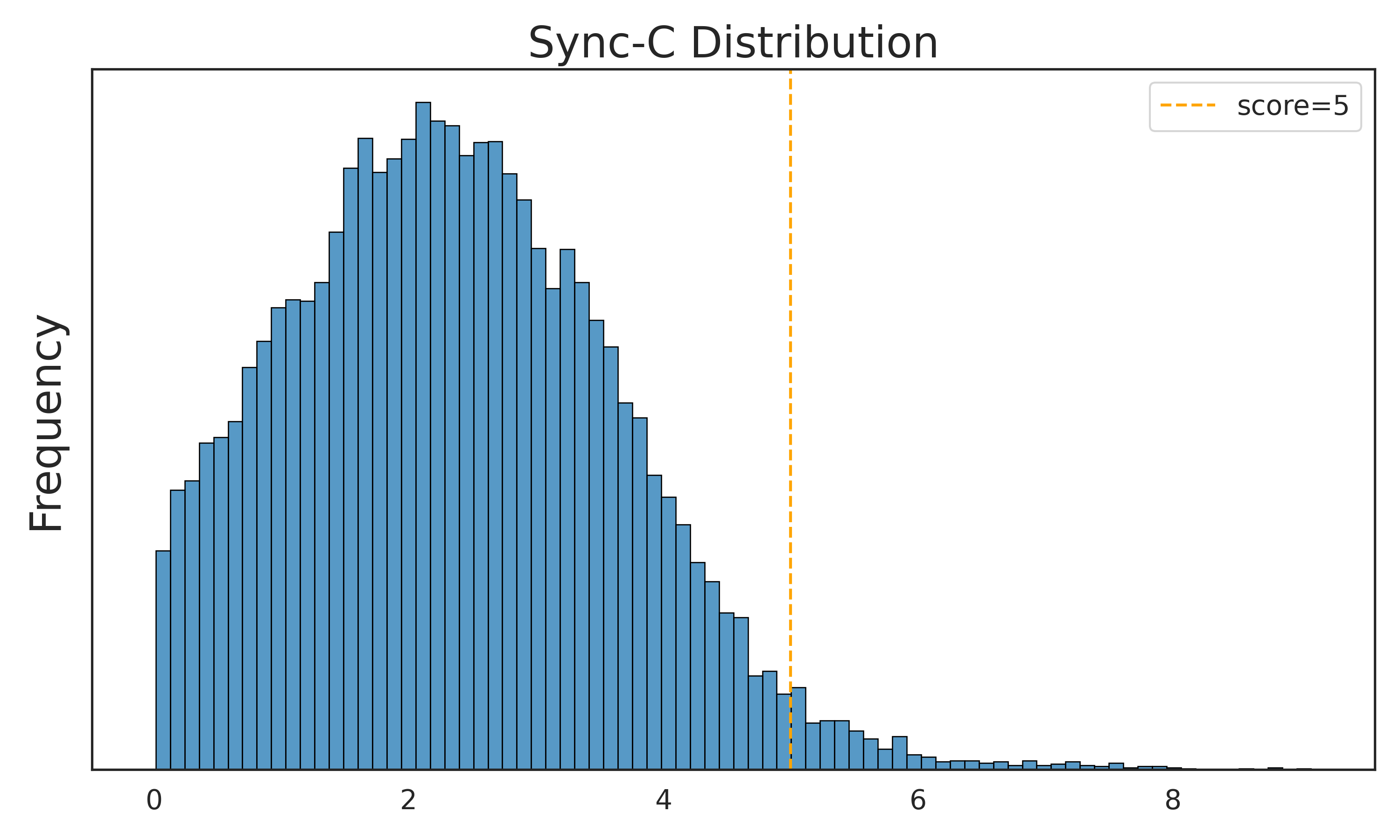}
    \vspace{-0.1in}
    \caption{Distribution of the Sync-C in the CelebV-HQ.}
    \label{fig:sync_c_dist} 
    \vspace{-0.1in}
\end{figure}

To further obtain high-quality talking video data, we developed a dedicated data processing pipeline for talking head generation. 
The pipeline consists of five steps:  

$\bullet$  First, we perform scene transition detection based on TransNet V2~\cite{soucek2024transnet} and trim video clips to a length of less than 30 seconds. 

$\bullet$ Second, we apply face detection based on Grounding DINO~\cite{liu2023grounding}, filtering out videos with no faces, partial faces, or multiple heads, and use the resulting bounding boxes to extract talking heads. To ensure that the cropped areas encompass more than just the human faces, we apply a scaling factor 1.1 to the bounding box regions.

$\bullet$ Third, we use HyperIQA~\citep{Su_2020_CVPR}, an image quality assessment model, to filter out low-quality and low-resolution videos. 
We apply HyperIQA to the first frame of each video and find that when the IQA score exceeds 40, there is a noticeable improvement in overall video quality. Therefore, we use an IQA score of 40 as a selection threshold, but this threshold can be dynamically adjusted depending on the dataset quality requirements.

$\bullet$ Fourth, we utilize SyncNet~\citep{prajwal2020lip} to filter out videos with audio-lip synchronization issues. The SyncNet Confidence (Sync-C) metric is used as the basis for filtering. Figure~\ref{fig:sync_c_dist} illustrates the confidence distribution of the CelebV-HQ dataset, where a threshold of 5 is applied for filtering. This threshold, like others, can be dynamically adjusted based on the dataset quality requirements.

$\bullet$ Lastly, we manually check the audio-lip synchronization and overall video quality for more accurate filtering for a subset of the data. After completing the entire pipeline, the total duration of the processed high-quality videos is about 660 hours.

\section{More Ablation Studies}\label{app:experiments}

\begin{figure}[t] 
    \centering
    \includegraphics[width=0.48\textwidth]{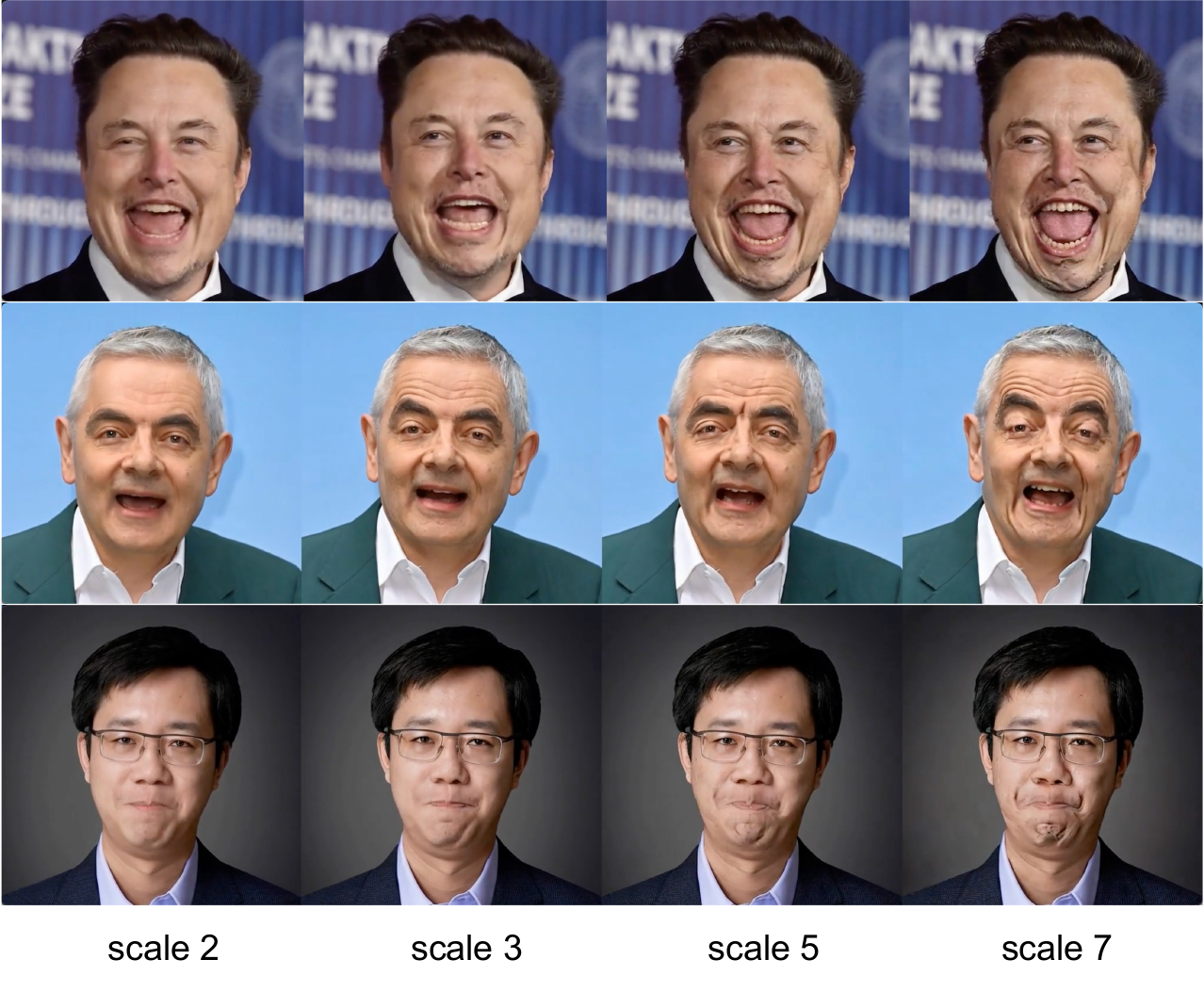}
    \vspace{-0.2in}
    \caption{Ablation of the classifier-free guidance scale. Please refer to the supplementary for video demos.}
    \label{fig:ablation_emo} 
\end{figure}

\paragraph{Classifier-free guidance scale.} By adjusting the classifier-free guidance scale, we observe variations in the expressiveness of the generated faces. As shown in Figure~\ref{fig:ablation_emo}, higher guidance scales lead to more pronounced emotional expressions. 

\section{More Qualitative Results}\label{app:qualitative}

\begin{figure}[t] 
   \centering 
   \includegraphics[width=0.42\textwidth]{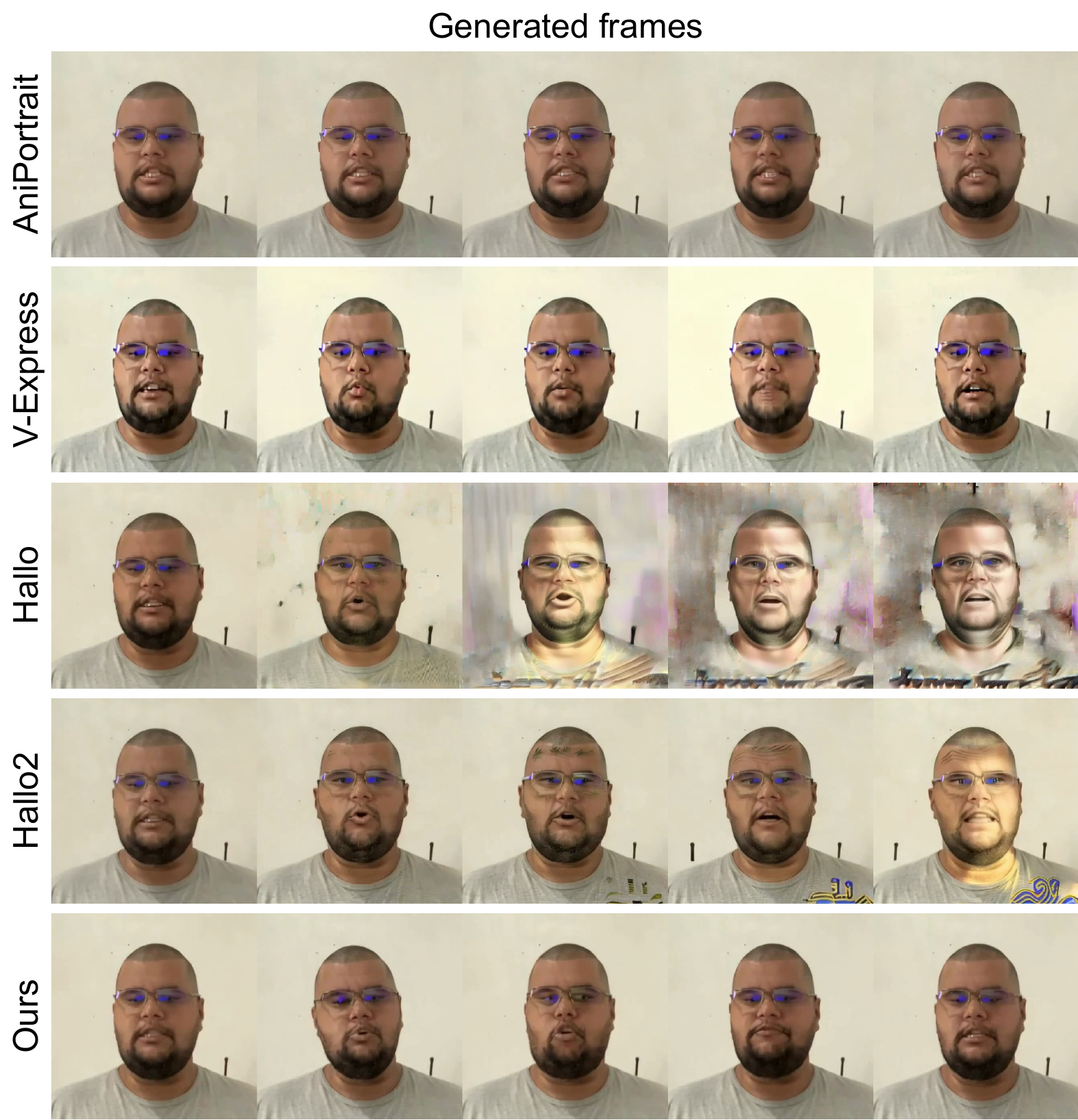}
   \caption{Visualization of generated videos on the OOD dataset. Existing methods either have poor audio-lip synchronization (\emph{e.g.}, AniPortrait~\citep{wei2024aniportrait}) or suffer from error accumulation (\emph{e.g.}, Hallo~\citep{xu2024hallo}). In contrast, \name generates talking videos with natural head motion and accurate audio-lip synchronization without artifacts. Please refer to the supplementary for video demos.}
   \label{fig:comparison} 
\end{figure}

\paragraph{Comparisons with baselines.} Figure \ref{fig:comparison} showcases comparisons between talking videos generated by \name and baseline models on sampled out-of-distribution (OOD) data. Specifically, some baseline models (\eg, Hallo and Hallo2) tend to produce artifacts and fail to preserve the original identity and fine details.  While certain methods (\eg, AniPortrait and V-Express) generate videos with fewer artifacts, they suffer from poor audio-lip synchronization and motion smoothness. In contrast, our method demonstrates the ability to produce more natural facial expressions and head movements that are well-aligned with the audio input. Additionally, the videos generated by \name exhibit superior overall visual quality and stronger identity consistency.

\begin{figure}[t]  
    \centering
    \includegraphics[width=0.42\textwidth]{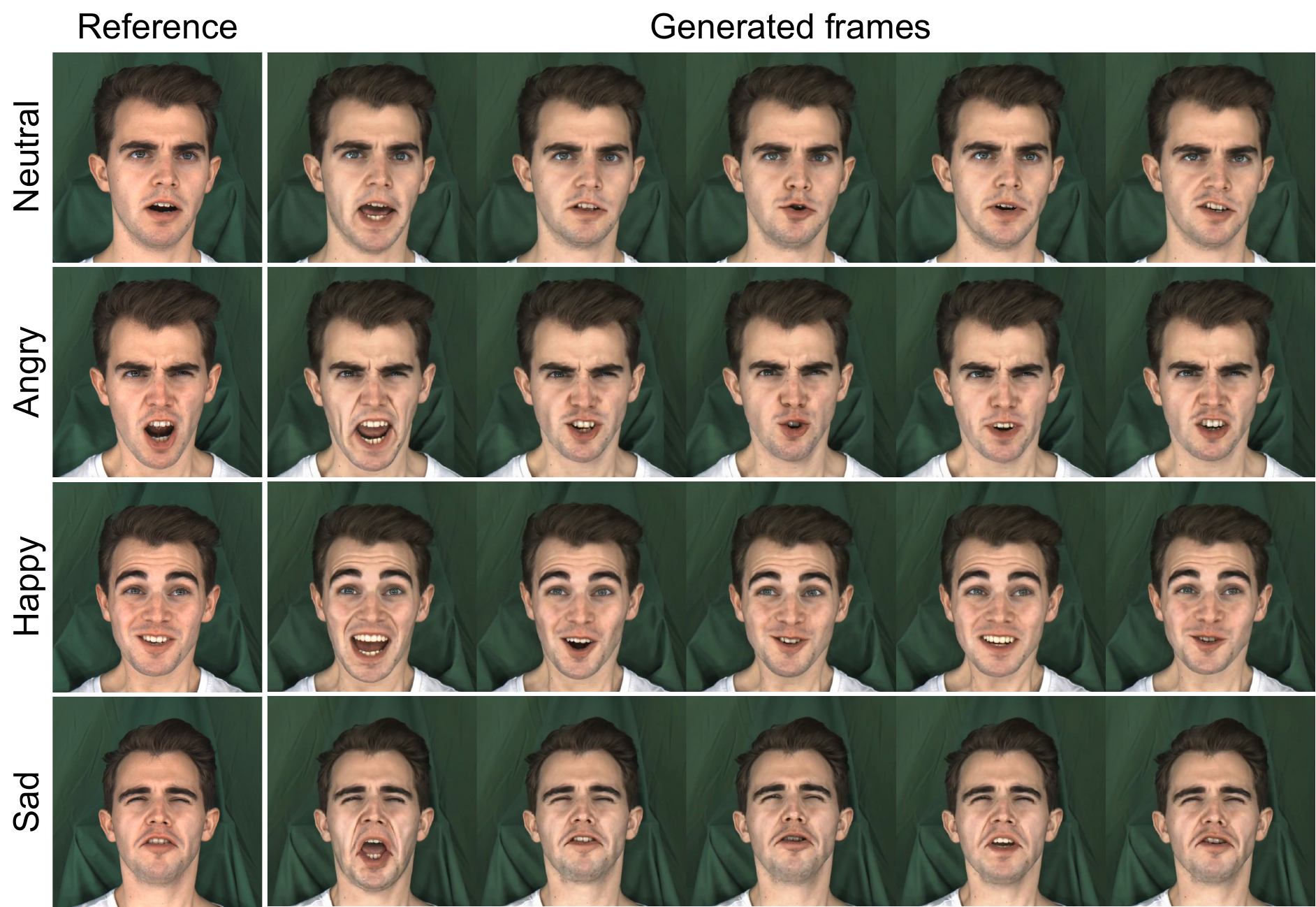}
    \caption{More visualization of expressive talking videos generated by MEMO based on reference images with various emotions. Please refer to the supplementary for video demos.}
    \label{fig:emotion_demo_diff_ref} 
\end{figure}

\begin{figure}[t]  
    \centering
    \includegraphics[width=0.42\textwidth]{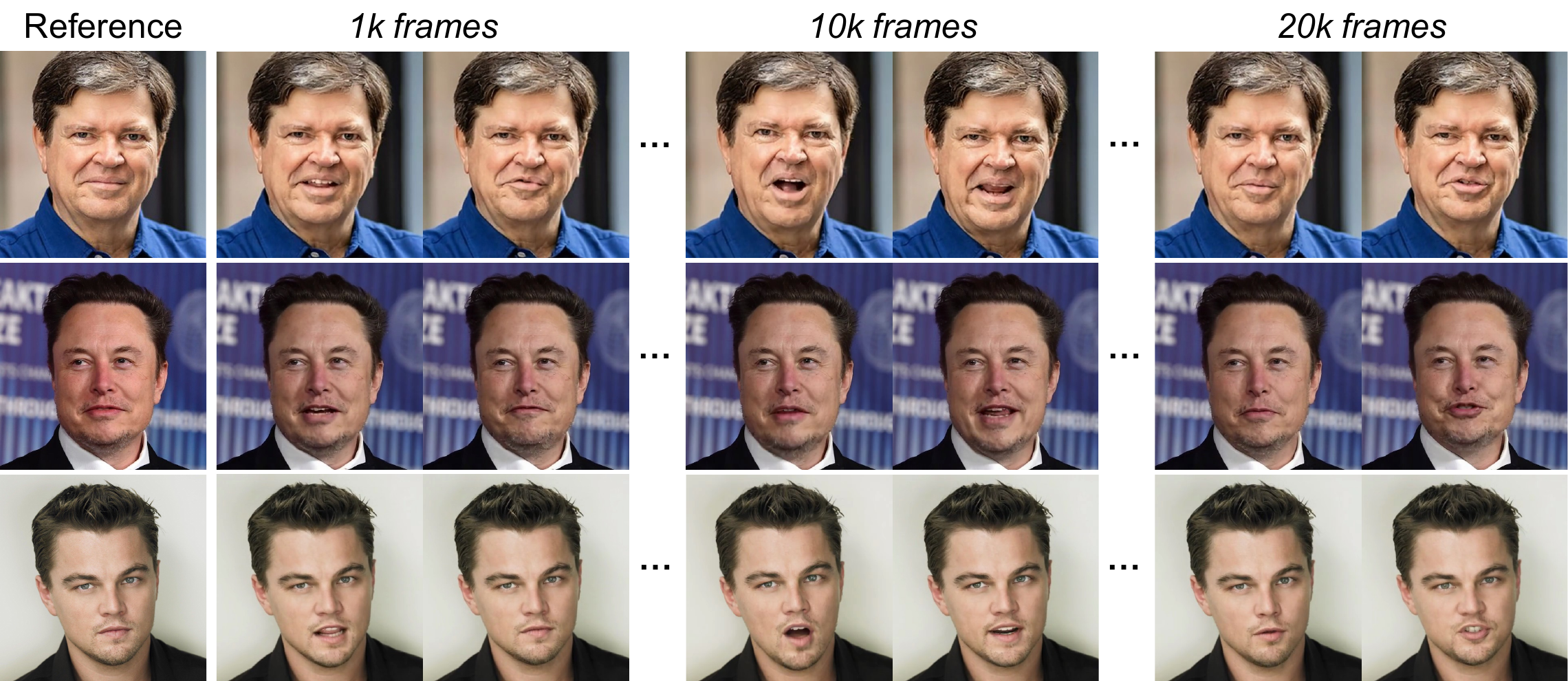}
    \caption{\name can generate long-duration videos with alleviated error accumulation and maintain identity consistency. Please refer to the supplementary for video demos.}
    \label{fig:long_demo} 
\end{figure}

\paragraph{More visualization of emotion-guided generation.} The facial expressions of the generated talking video are influenced by both the expressions in the reference image and the emotional tone of the audio. As discussed in Section~\ref{sec:audiomodule}, the overall emotional tone of facial expressions is inferred mainly from the facial expression of the reference image, while our audio emotion-aware module functions mainly as a subtle adjustment to enhance or moderately alter the emotion when prompted by the audio. Figure~\ref{fig:emotion_demo} in Section~\ref{sec_ablation} has demonstrated that given a fixed reference image, MEMO can refine the facial expressions of talking videos based on the given audio emotion. 

In this appendix, we further explore the flexibility of our method by evaluating its ability to generate expressive talking videos using reference images depicting the same person with different emotional expressions, such as neutral, angry, happy, and sad. To isolate the effect of reference image expressions, we set the audio emotion label to match the emotional state of each reference expression. As shown in Figure~\ref{fig:emotion_demo_diff_ref}, our method adapts seamlessly to diverse emotional states, generating highly expressive and emotionally consistent talking videos. These results highlight the robustness and versatility of \name in leveraging both reference expressions and audio cues to create emotionally nuanced talking videos.

\paragraph{Long-duration talking video generation.}  Figure~\ref{fig:long_demo} demonstrates that \name can generate long-duration videos while consistently maintaining the subject's facial features and expression fidelity over thousands of frames. The resulting videos exhibit smooth motion and high temporal coherence, showcasing the robustness of our approach for long video synthesis. These capabilities make \name a superior choice over existing methods for applications requiring extended video content with stable and consistent output quality.